\documentclass[lettersize,journal]{IEEEtran}
\usepackage{amsmath,amsfonts}
\usepackage{algorithmic}
\usepackage{array}
\usepackage[caption=false,font=normalsize,labelfont=sf,textfont=sf]{subfig}
\usepackage{textcomp}
\usepackage{stfloats}
\usepackage{url}
\usepackage{verbatim}
\usepackage{graphicx}
\def\BibTeX{{\rm B\kern-.05em{\sc i\kern-.025em b}\kern-.08em
    T\kern-.1667em\lower.7ex\hbox{E}\kern-.125emX}}
\usepackage{balance}

\usepackage{pifont}
\usepackage{booktabs}
\usepackage{color}
\usepackage{float}
\usepackage{enumerate}
\usepackage[linesnumbered,ruled,vlined]{algorithm2e}
\usepackage[colorlinks,linkcolor=green,anchorcolor=blue,citecolor=blue]{hyperref}
\usepackage[table]{xcolor}
\usepackage{soul}

\begin{document}
\title{Exploring Hyperspectral Anomaly Detection with Human Vision: A Small Target Aware Detector}
\author{Jitao Ma, Weiying Xie, \textit{Senior Member, IEEE}, Yunsong Li, \textit{Member, IEEE}
\thanks{J. Ma, W. Xie and Y. Li are with the State Key Laboratory of Integrated Services Networks, Xidian University, Xi’an 710071, China (e-mail: 21011210271@stu.xidian.edu.cn; wyxie@xidian.edu.cn; ysli@mail.xidian.edu.cn).

This work was supported in part by the National Natural Science Foundation of China under Grant 62121001, Grant 62322117, Grant 62371365, and Grant U22B2014; and in part by the Young Elite Scientist
Sponsorship Program through the China Association for Science and Technology under Grant 2020QNRC001.
}}

\markboth{Submitted to **}%
{How to Use the IEEEtran \LaTeX \ Templates}

\maketitle

\begin{abstract}
Hyperspectral anomaly detection (HAD) aims to localize pixel points whose spectral features differ from the background. HAD is essential in scenarios of unknown or camouflaged target features, such as water quality monitoring, crop growth monitoring and camouflaged target detection, where prior information of targets is difficult to obtain. Existing HAD methods aim to objectively detect and distinguish background and anomalous spectra, which can be achieved almost effortlessly by human perception. However, the underlying processes of human visual perception are thought to be quite complex. In this paper, we analyze hyperspectral image (HSI) features under human visual perception, and transfer the solution process of HAD to the more robust feature space for the first time. Specifically, we propose a small target aware detector (STAD), which introduces saliency maps to capture HSI features closer to human visual perception. STAD not only extracts more anomalous representations, but also reduces the impact of low-confidence regions through a proposed small target filter (STF). Furthermore, considering the possibility of HAD algorithms being applied to edge devices, we propose a full connected network to convolutional network knowledge distillation strategy. It can learn the spectral and spatial features of the HSI while lightening the network. We train the network on the HAD100 training set and validate the proposed method on the HAD100 test set. Our method provides a new solution space for HAD that is closer to human visual perception with high confidence. Sufficient experiments on real HSI with multiple method comparisons demonstrate the excellent performance and unique potential of the proposed method. The code is available at \url{https://github.com/majitao-xd/STAD-HAD}.
\end{abstract}


\section{Introduction}
\IEEEPARstart{H}{yperspectral } images (HSIs) cover dozens or even hundreds of subdivided spectral bands in the electromagnetic spectrum, including information that is difficult for human vision to detect, and thus reflect the essential characteristics of different materials \cite{hsi}. With the development of spectral imagers, hyperspectral imaging has been widely used in various fields, such as crop growth detection \cite{nongye-hsi}, mineral exploration \cite{kuangchan-hsi}, disaster detection \cite{zaihai-hsi} and emergency rescue \cite{jiuyuan-hsi}. Hyperspectral image processing, including hyperspectral anomaly detection \cite{had}, hyperspectral classification \cite{hc, tcsvt2, tcsvt3, tcsvt4}, multi-modal fusion \cite{ldx, tcsvt5}, and hyperspectral target detection \cite{htd}, is key tasks in these applications. Among them, hyperspectral anomaly detection (HAD) is a vital task that provides pixel-level accurate localization of anomalous regions in the HSI. 

Anomalous regions in HSI typically include unknown materials and unanticipated objects, none of which have a prior knowledge. Importantly, the vast majority of anomalies obey two assumptions: (1) the anomalies are significantly different from the surrounding background both spatially and spectrally, and (2) the anomalies are small targets. These are also the core ideas of the current HAD. Based on these two assumptions, a series of HAD methods have been proposed, and they can be mainly grouped into the traditional methods and deep learning-based methods.

Traditional methods are devoted to mathematically analyzing the differences between spectral vectors as a way of distinguishing anomalies from the background. RX \cite{rx} assumes that the background obeys a normal distribution and takes the global Mahalanobis distance between spectral vectors as the anomaly score. On this basis, LRX \cite{lrx} builds a background model with kernel density estimation, and design a concentric dual rectangular window centered on the sample to be measured to achieve local hyperspectral anomaly detection. Inspired by this, several dual-window based approaches have been proposed, such as QLRX \cite{qlrx} and LAIRX \cite{lairx}. In order to avoid the contamination of background statistics by anomalous data, representation-based background modeling methods have been proposed. CRD \cite{crd} assumes that the background spectral vector can be linearly approximated by the spectral vectors around it, and builds a background model by this decision. SWCRD \cite{swcrd} further considers the importance of different spectral bands. Different from CRD based on collaborative representation, SRD \cite{srd} is based on sparse representation, which assumes that each sample in the HSI can be represented by several elements in a dictionary of hyperspectral features. LRASR \cite{lrasr} discovered that the coefficients of the representation of the background spectral vectors can form a low-rank matrix and introduced the low-rank representation to model the background. LSDM-MoG \cite{lsdm} employs a hybrid noise model on this basis to more accurately represent complex distributions. In addition, in order to utilize the rich spatial information of HSI, a series of HAD methods based on background removal have been proposed. AED \cite{aed} combines attribute filters and differential operations to remove the background. STDG \cite{stdg} introduces structure tensor and guided filtering. Further, the tensor-based method can explore the 3D structure of the HSI and extract the anomaly components by tensor decomposition. TDAD \cite{tdad} distinguishes between background and anomaly by estimating three factor matrices and a core tensor via the Tucker decomposition. TPCA \cite{tpca} applies principal component analysis in separating background and anomalies. PTA \cite{pta} utilizes the nuclear norm and $l_{1,2}$-norm to process the background and anomaly parts, respectively, and obtains the anomaly through the anomaly component. Recently, there have been some approaches to HAD that provide novel ideas from other perspectives. FrFE \cite{frfe} applies fractional-order Fourier transform as pre-processing to obtain features and introduces the Shannon entropy uncertainty principle to significantly distinguish the signal from background and noise. SSDF \cite{ssdf} combines dimensionality reduction and data splitting techniques with an isolation-based discriminative forest model designed to avoid background contamination caused by anomalous targets. However, these traditional methods are model-driven based and require a lot of experience to set up the parameters. 

Recently, deep learning-based methods are popular in HAD. Deep learning-based approaches are primarily data-driven, learning the distribution of HSI to separate anomalies from backgrounds. Due to the scarcity of anomalous samples, a series of deep learning-based HAD methods are unsupervised. With the above two assumptions, the neural network fits the distribution of the background spectral vectors more easily, with a small error before and after the reconstruction. On the contrary, the error before and after anomalous reconstruction is large. Therefore, reconstruction errors are commonly used as anomaly score maps in unsupervised HAD methods. Especially, some generative models such as AE \cite{ae}, AAE \cite{aae}, GAN \cite{gan} have emerged as the building blocks of unsupervised HAD methods. SAFL \cite{safl} takes AAE as backbone and introduces spectral constraint loss to reduce the false alarm rate. SDLR \cite{sdlr} further adds the spectral angular distance to the AAE. HADGAN \cite{hadgan} proposes spectrally constrained GAN networks to achieve stronger generative performance. These methods use fully connected layers as the base composition of the neural network, which makes it difficult to learn the spatial features of HSIs. Therefore, to improve detection accuracy, several methods introduce additional spatial information. Transferred deep CNN \cite{tdcnn} obtains the differences between pixel pairs by convolution and classifies the results. Auto-AD \cite{autoad} designed a fully convolutional AE and adaptive weighted loss function to obtain high contrast anomaly score maps. RGAE \cite{rgae} embeds a superpixel segmentation-based graph regularization term into AE to maintain the geometric structure and local spatial consistency of HSIs. In order to utilize the low-dimensional manifolds of HSI more effectively, a series of unsupervised HAD methods based on density estimation have been proposed. E2E-LIADE \cite{e2e} takes AE as a reconstruction network and fuses the reconstruction error space and extracted features into a density estimation network fitting a Gaussian mixing model. LREN \cite{lren} further considers the physically mixing properties in hyperspectral imaging and constructs characterization dictionaries to better fit the Gaussian mixing model. However, these methods assume HSIs obey a specific distribution that is not applicable to other scenarios. Despite not requiring additional labeling and complex tuning, unsupervised HAD methods require retraining in new test scenarios which limits their application to edge devices with high real-time demands. While semi-supervised and self-supervised HAD methods are effective tools to solve this problem while utilizing the abundant anomaly-free hyperspectral data. WeaklyAD \cite{wad} obtains pseudo-labeling through coarse detection to guide subsequent weakly supervised learning. The DFAE \cite{dfae} decomposes the input HSIs into high and low frequency components and trains them separately using two AEs. AETNet \cite{aetnet} allows the network to learn contextual features of background and anomalies by adding randomly generated pseudo-anomalies to the input HSIs. All three of the above methods achieve the generalization of the trained network to the test set without the need for additional training. However, weaklyAD relies on the performance of coarse detection and is susceptible to mislabeled noise, while DFAE adopts too many hyperparameters during post-processing. The data enhancement approach of pseudo anomalies in AETNet, on the other hand, does not take into account the different criteria for anomalies in different classes, which possibly lead to biased training distributions. Besides, the large size of AETNet makes it unsuitable for use with edge devices such as satellites.

\begin{figure}[t]
	\centering
	\includegraphics[width=0.5\textwidth]{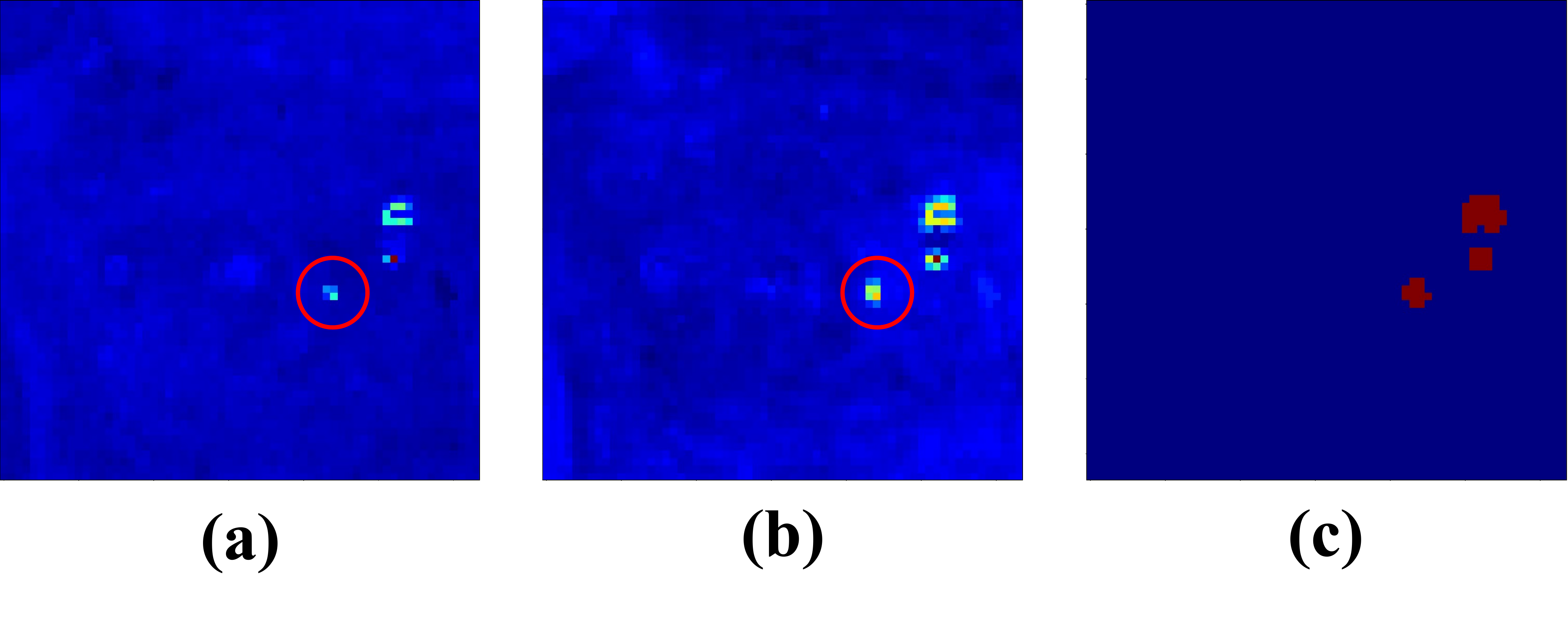}
	\caption{One of the results of testing on the HAD100 dataset. (a) Result of reconstruction-based method, the anomalies within the red circles are not detected because of their easy reconstruction. (b) Results of the saliency map based method, more anomalies are detected compared to reconstruction-based method. (c) Anomalies map.}
	\label{fig:1}
\end{figure}

In summary, HAD requires a parameter-less, lightweight and effective self-supervision approach. Following the two assumptions of anomalous targets, we propose a small target aware detector (STAD) for self-supervised HAD. STAD is based on deep learning without pre-processing and post-processing, so as to avoid overparameterization. Different from existing major deep learning based approaches, we do not directly take the output of the neural network as anomaly score. Instead, we explore the feature space of HSI based on deep learning. The reason lies in the possibility of failure in other solution spaces, such as the commonly used reconstruction error space. Reconstruction errors are strongly influenced by the distribution of the data, especially the anomalous spectral vectors that are easy to reconstruct, which possibly leads to low confidence results, as shown in Fig. \ref{fig:1} (a). The essential reason for this problem is that the reconstruction error space does not correspond to human visual perception.

Lately, Zhang et al. \cite{feature} prove that feature space is the firm perceptual representation of the input image, which corresponds more to human visual perception. However, the feature maps of HSI cannot directly separate the anomalies from the background and contain noise. For instance, even though methods like LREN extract HSI features, they still require estimating the background distribution, and their fundamental solution process remains in the probability density space. Therefore, in order to obtain maps with richer anomaly representations and corresponding more to human visual perception in the feature space for direct separation of anomalies from the background, we introduce the saliency map \cite{saliency}. 

Saliency map is commonly applied for feature map visualization and neural network model interpretation. It is the derivative of the feature map with respect to the loss function. In particular, we back-propagate loss of the network to the input HSI so that we transfer the solution process to the feature space, which corresponds more to human visual perception and contains more anomalous representations, as shown in Fig. \ref{fig:1} (b). Besides, the saliency map can also respond to the region that the neural network is interested in the input image. Thus, the salience map can be applied to any network architecture. It is worth noting that we only use the reconfigured network in this paper, which represents that there is still much to be explored in HAD based on feature space, and we discuss that specifically in Section \ref{method}. Further, to localize small targets, we design a small target filter (STF) which combines the global Mahalanobis distance and the bilateral filter. It can reduce the detection false alarm rate by decreasing the anomaly score of regions with large size and low confidence level. 

Considering the domain-specific requirements for real-time HAD, and the fact that self-supervised HAD methods require a large size of models to fit numerous normal samples, we build a complex teacher network and a simple student network, and distill the knowledge learned in the teacher network to the student network. For neural network design, we propose a knowledge distillation strategy of fully connected network to convolutional network. We use Transformer with multiple fully connected layers as backbone of the teacher network, which can learn the contextual features of the input HSIs while avoiding the damage of deeply layered CNNs to the edge features. Instead, the student network is a simple convolutional AE. In addition to lightening the model, this strategy allows the student network to learn what the teacher network has learned about spectral information while adding additional spatial information, which can improve detection performance. During the training process, we take numerous anomaly-free samples as a training set for the teacher network and distill the trained teacher network to the student network. During the testing process, we locate small anomaly targets by STAD, and get the anomaly score map. We verify the proposed method on the HAD100 \cite{aetnet} dataset, and our method has the best performance compared to several representative HAD methods. The contributions of this article can be summarized as follows.

\begin{itemize}
	\item For the first time, we transfer the solution of the HAD problem to the feature space, which corresponds more to human visual perception and contains more anomalous representations, and opens up a new solution space.
	\item To obtain a more accurate anomaly score map in the feature space, a small target aware detector is designed for HAD that focuses the region of interest of the network on the input HSI to small targets, and reduces the impact of low confidence regions on the detection results.
	\item In order to satisfy the high real-time requirements in some scenarios, a knowledge distillation strategy from fully connected network to convolutional network is designed, which reduces the model size while preserving spectral and spatial features.
	\item Sufficient experiments with several representative comparison methods on the HAD100 dataset demonstrate that the proposed method performs optimally with the highest average accuracy and average background suppressibility. In addition, ablation study and dependency study demonstrate the unique potential of feature space-based approaches in HAD.
\end{itemize}

\section{Related work}
This section briefly describes two key techniques used in this paper. These include saliency map and knowledge distillation, both of which are commonly used in natural image processing, and the latter is also widely used in industrial anomaly detection.

\subsection{Saliency Map}
\label{2a}
Saliency map for neural networks is first used to visualize and understand CNNs \cite{saliency}. It can extract spatial representation information from a specified class with a single backpropagation and can be used for weakly supervised object localization. Different from common backpropagation, this method restricts the activation of neurons. Briefly, only one of the classes in the classification result is selected for backpropagation, usually choosing the most confident classification result of the network, or the true classification in the labels. Similarly, Jost et al. \cite{gbp} proposes guided backpropagation, in which only neurons with inputs and gradients greater than 0 are activated in the ReLu layer. 

Zhou et al. \cite{cam} study a different aspect of this, proposing class activation mapping (CAM). CAM locates the position of objects that are of interest to the network through global average pooling. However, the method requires the model to have a global average pooling layer. On this basis, Selvaraju incorporates backpropagation and proposes the Grad-CAM \cite{grad-cam}. It obtains the saliency map by backpropagation, then computes the saliency map as a channel attention vector and multiplies it to the feature map. Although Grad-CAM can be applied to different network structures, it ignores the contribution of each element in the saliency map. Therefore, Chattopadhay et al. \cite{grad-campp} proposed Grad-CAM++, which adds additional weights to each element on the saliency map. Besides, Wang et al. \cite{score-cam} argue that the possible noise in the gradient can lead to erroneous results in Grad-CAM-based methods, and propose Score-CAM which requires no gradient. The same authors then proposed SS-CAM \cite{ss-cam}, which utilizes a smoothing strategy to reduce noise. 

In this paper, we obtain saliency maps based on simple backpropagation. Meanwhile, considering that the reconstruction-based HAD method obtains not classification results but anomaly scores, we use a mask matrix to shield the backpropagation of low-confidence results. The mask matrix is obtained by STF and thus also serves to localize small targets. 

\subsection{Knowledge Distillation}
Knowledge distillation consists of two networks, the teacher network and the student network. Typically, a trained teacher model provides the knowledge and the student model is trained to acquire the teacher's knowledge through distillation \cite{kd}. It allows for the migration of knowledge from a complex teacher model to a simple student model at the cost of a slight performance loss. 

As a lightweight method, knowledge distillation can significantly reduce the computational cost during testing. However, how the distillation loss function student network are designed is critical to ensure performance. Hinton et al. \cite{kd} defined distillation loss as minimizing the logits of teacher and student networks. Ba et al. \cite{kd2} replace logits with probability values. Romero et al. \cite{kd3} utilized not only the logits of the teacher network, but also its intermediate parameters. Komodakis et al. \cite{kd4} propose knowledge distillation using the attention. Yim et al. \cite{kd5} consider the relationship between layers of the teacher network. Wang et al. \cite{kd6} design a special knowledge distillation method for GANs. And Lin et al. \cite{kd7} design knowledge distillation method for Transformer. 

The above methods provide illuminating ideas on distillation loss, but are based on the structure of convolutional layers in designing both the teacher network and the student network. Due to the low spatial resolution and high spectral resolution of HSI, deeply layered CNNs may compromise the edge features of the image. In this article, we design the teacher network and student network as fully connected and convolutional networks, respectively. This strategy allows the teacher network to learn the data distribution and spectral characteristics of numerous HSIs, yet adds additional spatial information to the student network. Thus, both generalizability and performance can be guaranteed.

\section{Method}
\label{method}
In this section, we present the details of the proposed small target-aware detector. We first analyze the motivation and idea of solving the HAD problem in feature space, then we present the details of our approach, and finally we discuss further research on feature space-based HAD methods.

\subsection{Feature Space}
The evolution of neural networks has consistently aimed to emulate human visual perception. However, human perception is influenced by various factors, including higher-order image structure, contextual information, and subjective judgment. This variability is why methods solely relying on reconstruction errors may encounter challenges; they do not align with the intricacies of human visual perception.

The reconstruction error-based Hyperspectral Anomaly Detection (HAD) approach operates on the assumption that difficult-to-reconstruct vectors are anomalies, while easy-to-reconstruct vectors are considered background. However, this assumption does not always hold true in real datasets. Real scenarios may include anomaly vectors that are easy to reconstruct and background vectors that are difficult to reconstruct. While this is a straightforward issue for human visual perception, it introduces a critical flaw in the reconstruction error space. Interestingly, the computer vision community has discovered that feature maps generated by neural networks serve as an excellent representation space and correspond closely to human visual perception \cite{feature}. This discovery inspires us to address the HAD problem by leveraging insights from human visual perception and the feature space.

\begin{figure}[t]
	\centering
	\includegraphics[width=0.5\textwidth]{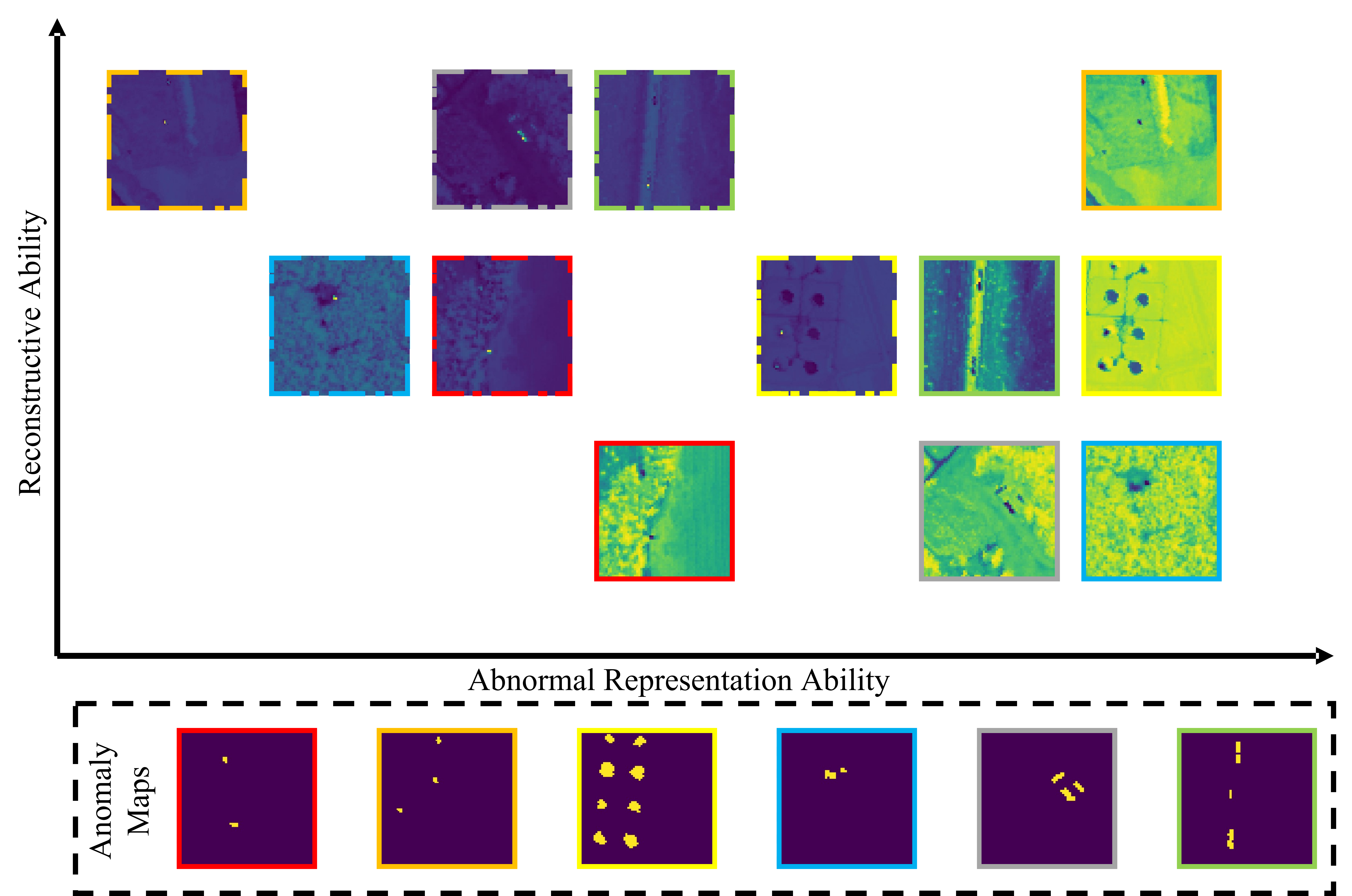}
	\caption{Schematic representation of the anomalous representation ability of feature maps and reconstruction error maps. A set of maps with the same color border represents the label (bottom row), feature map (solid box), and reconstruction error map (dashed box) of the same HSI. Note that the feature maps are obtained by solving for the maximum value of the channel dimension, which does not imply that the detection results are so. The reconstruction errors are calculated by mean square error (MSE).}
	\label{fig:2}
\end{figure}

Building on the aforementioned motivation, we analyze the feature space and reconstruction error space of HSI. Specifically, we demonstrate the capability of feature maps and reconstruction error maps to represent anomalous targets in Fig. \ref{fig:2}. It is evident that the reconstruction performance does not adequately reflect the ability to represent anomalous targets, whereas the feature map excels in capturing the location and shape of anomalies. Additionally, the feature map exhibits superior robustness and performs well across various scenarios. Therefore, transferring the solution of the HAD problem from the reconstruction error space to the feature space becomes essential.

In summary, feature space is more advantageous for anomalous target representation. Further, we define the core problem as how to obtain a feature map that corresponds more to human visual perception. Simple methods, such as directly taking the response of the feature map in the channel dimension or employing principal component analysis, introduce significant noise. Therefore, we apply feature map visualization methods specifically designed for human vision. The saliency map is just one simple example of these methods, but it is sufficient to validate our ideas.

\subsection{Small Target Aware Detector}
\subsubsection{Saliency Map}
In this subsection, we present a common method for computing saliency maps. Given a test HSI $\boldsymbol{H}=\{\boldsymbol{h}_1,\ldots,\boldsymbol{h}_i,\ldots,\boldsymbol{h}_L\}\in \mathbb{R}^{L \times B}$, and a trained network $f(\cdot)$, where $\boldsymbol{h}_i$, $(i=1,2, \ldots, L)$ denotes the $i$-th spectral vector with $B$ bands, $L$ denotes the total number of spectral vectors, and for HSI with height $M$ and width $N$, respectively, $L=M\times N$. We feed $\boldsymbol{H}$ into the network to obtain the output by

\begin{equation}
	\label{e:1}
	\mathcal{L}=f(\boldsymbol{H}).
\end{equation}

Subsequently, we obtain the gradients of HSI by

\begin{equation}
	\label{e:2}
	\boldsymbol{g}_i = \frac{\partial \mathcal{L}}{\partial \boldsymbol{h}_i}
\end{equation}
where $\boldsymbol{g}_i$ and $\boldsymbol{h}_i$ have the same shape. The element corresponding to $\boldsymbol{h}_i$ in the saliency map is the maximum response of $\boldsymbol{g}_i$, and calculated according to

\begin{equation}
	\label{e:3}
	s_i=\max(|\boldsymbol{g}_i|),
\end{equation}
where $|\cdot|$ denotes absolute value operation. Finally, we transform $S=\{s_1,\ldots,s_i,\ldots,s_L\}\in\mathbb{R}^{L \times 1}$ into the saliency map of size $M \times N$.

\begin{figure*}[!t]
	\centering
	\includegraphics[width=0.95\textwidth]{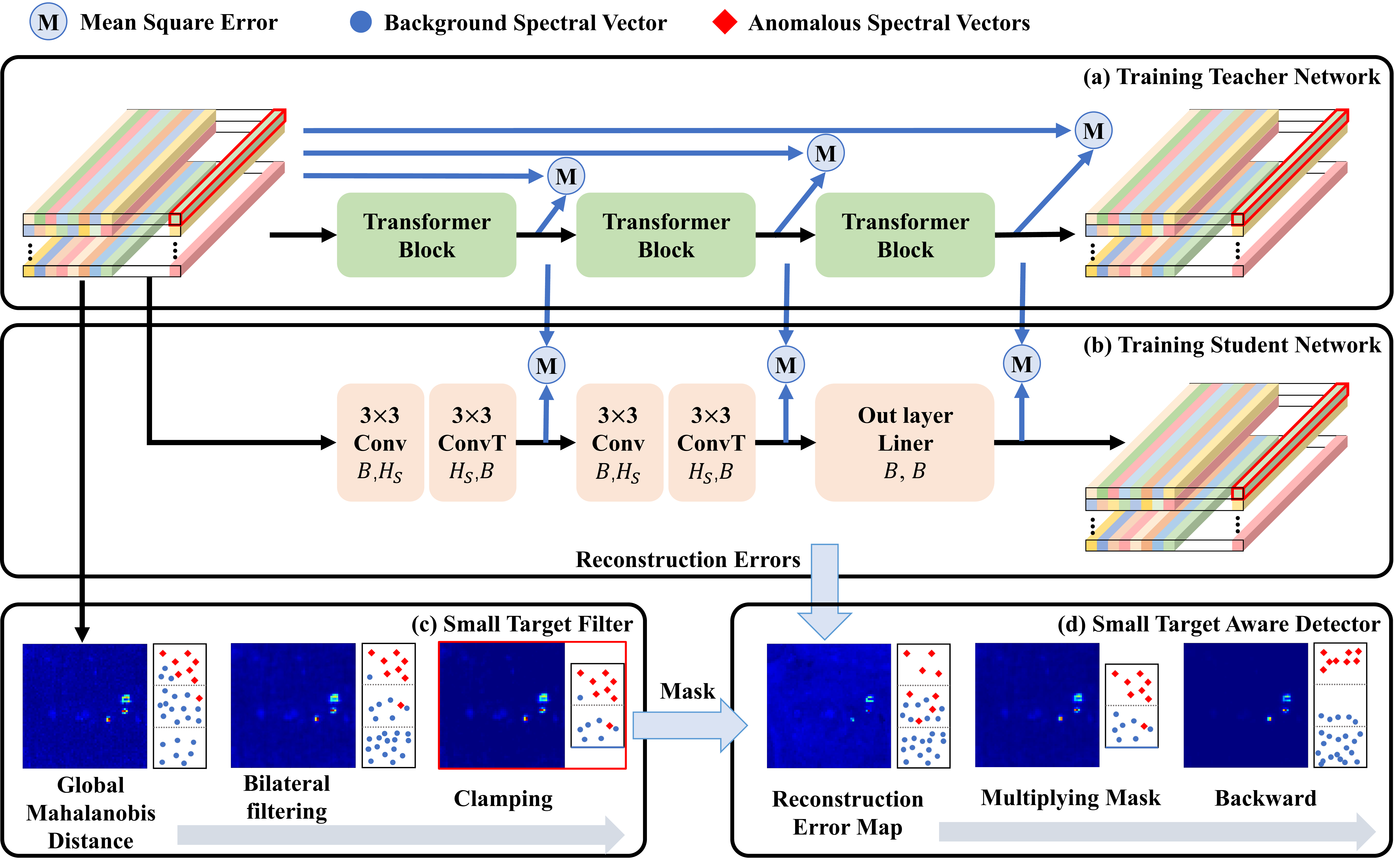}
	\caption{Schematic of our proposed approach. (a) The training process of the teacher network. (b) The process of training student network through knowledge distillation. (c) Obtaining the mask matrix through STF. (d) Detecting and localizing anomalous targets in HSI by STAD.}
	\label{fig:4}
\end{figure*}

\subsubsection{Teacher Network Training Process}
As described by the equations in the previous section, obtaining a saliency map necessitates a well-trained neural network. Consequently, we introduce our self-supervised network and knowledge distillation strategy in this section.

\begin{figure}[t]
	\centering
	\includegraphics[width=0.5\textwidth]{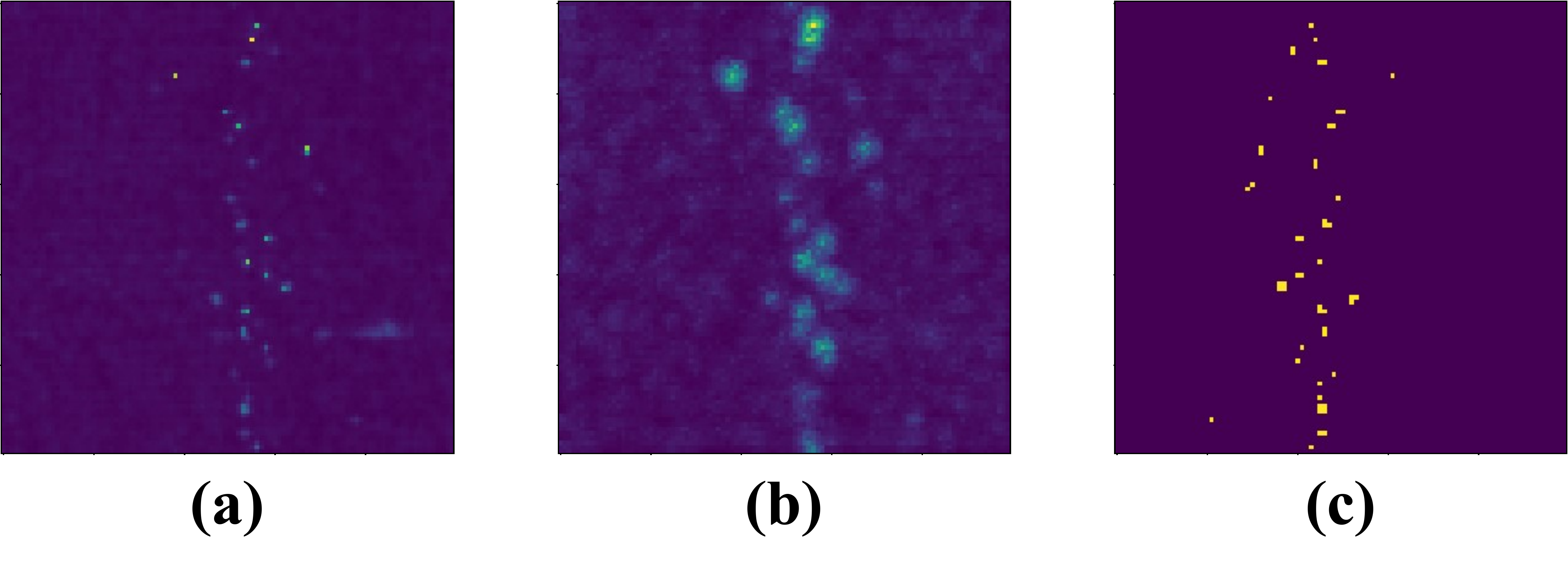}
	\caption{Feature maps of the intermediate layers of the fully connected layer network (a) and the convolutional layer network (b), and (c) is the anomaly map.}
	\label{fig:3}
\end{figure}

Due to the low resolution of HSIs, the edges between neighboring pixels lack smoothness. Consequently, multilayer convolution compromises the edge features of HSIs, as shown in Fig. \ref{fig:3}. While the teacher network requires an adequate number of network layers to fit the data distribution effectively, leading us to choose the fully connected network architecture. In the network architecture with a fully connected layer as the base layer, the Transformer exhibits the best performance. Thus, we design teacher networks with Transformer architecture, as shown in Fig. \ref{fig:4}. The teacher network consists of three Transformer blocks, each with two fully connected layers. The input to the teacher network is a HSI training set $\boldsymbol{D}_{train}$ without anomalies. The outputs of the teacher network are

\begin{equation}
	\label{e:4}
	\boldsymbol{\hat{D}}_{t1},\boldsymbol{\hat{D}}_{t2},\boldsymbol{\hat{D}}_{t3}=f_t(\boldsymbol{D}_{train};\theta_t),
\end{equation}
where $\boldsymbol{\hat{D}}_{t1}$, $\boldsymbol{\hat{D}}_{t2}$ and $\boldsymbol{\hat{D}}_{t3}$ denote the reconstruction of $\boldsymbol{D}_{train}$ by different blocks of teacher network respectively. Here, $f_t(\cdot)$ denotes the teacher network, and $\theta_t$ represents the parameter of the teacher network. The loss function of the teacher network is defined as

\begin{equation}
	\label{e:5}
	\mathcal{L}_t=\left \| \boldsymbol{\hat{D}}_{t1}-\boldsymbol{D}_{train} \right \|_2 + \left \| \boldsymbol{\hat{D}}_{t2}-\boldsymbol{D}_{train} \right \|_2 + \left \| \boldsymbol{\hat{D}}_{t3}-\boldsymbol{D}_{train} \right \|_2
\end{equation}
where $\left \| \cdot \right \|_2$ denotes the $l_2$-norm. It is crucial to note that our teacher network is essentially a reconfiguration network, based on two considerations: 1) The training process is self-supervised. 2) As an implicit comparison, reconfiguration-based networks better highlight the advantages of feature space. 

\subsubsection{Student Network Training Process}
Due to the limited number of layers in the student network, the convolution operation does not compromise the features of HSIs. Instead, a small number of convolutional layers can introduce additional spatial information. Consequently, we combine convolutional and deconvolutional layers as blocks of the student network, and the last layer utilizes a fully connected layer, as shown in Fig. \ref{fig:4}. During the training phase, we still take $\boldsymbol{D}_{train}$ as input, then the outputs are

\begin{equation}
	\label{e:6}
	\boldsymbol{\hat{D}}_{s1},\boldsymbol{\hat{D}}_{s2},\boldsymbol{\hat{D}}_{s3}=f_s(\boldsymbol{D}_{train};\theta_s),
\end{equation}
where $\boldsymbol{\hat{D}}_{s1}$, $\boldsymbol{\hat{D}}_{s2}$ and $\boldsymbol{\hat{D}}_{s3}$ denote the reconstruction of $\boldsymbol{D}_{train}$ by different blocks of student network respectively, $f_s(\cdot)$ denotes the student network, and $\theta_s$ is the parameter of the student network. To distill the knowledge learned by the teacher network into the student network, we align the outputs of both networks. Simultaneously, as the feature space of the teacher network contains a rich representation of anomalies, we also align the intermediate layer outputs of these two networks. The loss function of the student network is defined as

\begin{equation}
	\label{e:7}
		\mathcal{L}_{kd}=\left \| \boldsymbol{\hat{D}}_{s1}-\boldsymbol{\hat{D}}_{t1} \right \|_2 + \left \| \boldsymbol{\hat{D}}_{s2}-\boldsymbol{\hat{D}}_{t2} \right \|_2 + \left \| \boldsymbol{\hat{D}}_{s3}-\boldsymbol{\hat{D}}_{t3} \right \|_2
\end{equation}

Finally, after numerous iterations, we can obtain a well-trained student network.

\subsubsection{Small Target Filter}
Before computing the saliency map, we need to mask low confidence neurons. In tasks such as natural image classification, this is typically accomplished by setting the error class to zero. However, in HAD, anomalies are detected as an anomaly score map, representing a grayscale map with the same number of pixels as the original HSI. The value of each pixel in this map indicates its degree of anomaly. Consequently, we mask the low-confidence results using a masking matrix. The mask matrix is obtained via Small Target Filter (STF), which operates under the assumption that anomalies resemble small targets. 

Given an input test HSI $\boldsymbol{H}=\{\boldsymbol{h}_1,\ldots,\boldsymbol{h}_i,\ldots,\boldsymbol{h}_L\}\in \mathbb{R}^{L\times B}$, STF first computes its global Mahalanobis distance, $\boldsymbol{Z}=\{z_1,\ldots,z_i,\ldots,z_L\}\in \mathbb{R}^{L\times 1}$by

\begin{equation}
	\label{e:8}
	z_i = {\rm sum} \left( \boldsymbol{\overline{h}} \times \boldsymbol{C}^{-1} \odot \boldsymbol{\overline{h}} \right)
\end{equation}
where $\boldsymbol{\overline{h}} = \boldsymbol{h}_i-\frac{1}{L}\sum_{j=1}^{L}\boldsymbol{h}_j$, ${\rm sum}(\cdot)$ denotes the sum of all elements in the vector, $\odot$ denotes the Hadamard product, $\boldsymbol{C}=\{c_{ij}|i,j=1,2,\ldots,B\}\in \mathbb{R}^{B\times B}$ denotes the covariance matrix of $\boldsymbol{H}$, $c_{ij}$ is calculated by

\begin{equation}
	\label{e:9}
	c_{ij}=\frac{1}{B-1}{\rm sum} \left( (\boldsymbol{h}_i-\frac{{\rm sum}(\boldsymbol{h}_i)}{B}) \odot (\boldsymbol{h}_j-\frac{{\rm sum}(\boldsymbol{h}_j)}{B}) \right).
\end{equation}

Transform $\boldsymbol{Z}$ into a matrix of size $M\times N$ to obtain the global Mahalanobis distance map $\boldsymbol{Z}=\{z_{ij}|i=1,2,\ldots,M;j=1,2,\ldots,N\}\in \mathbb{R}^{M\times N}$. While the global Mahalanobis distance map can effectively localize small targets, it also introduces noise. To mitigate the impact of noise, we employ a bilateral filter to process the global Mahalanobis distance map. The filter output $\hat{z}_{ij}$ is

\begin{equation}
	\label{e:10}
	\hat{z}_{ij}=\frac{\sum_{(k,l)\in {\rm A}(i,j;r)} z_{kl} w(i,j,k,l)}{\sum_{(k,l)\in {\rm A}(i,j;r)} w(i,j,k,l)},
\end{equation}
where ${\rm A}(i,j;r)$ denotes the pixels in the circle with $r$ as the radius and $d_{ij}$ as the center point, $w(i,j,k,l)$ denotes the weight factor, calculated according to

\begin{equation}
	\label{e:11}
	w(i,j,k,l)=\exp^{-\frac{(i-k)^2+(j-l)^2}{2\sigma^2_s} - \frac{\left|z_{ij}-z_{kl}\right|^2}{2\sigma^2_c}},
\end{equation}
where $\sigma_s$ and $\sigma_c$ denote the standard deviation of the Gaussian filtering kernels in the spatial and value domains, respectively. Finally, we get the small target mask matrix $\boldsymbol{\hat{Z}}=\{\hat{z}_{ij}|i=1,2,\ldots,M;j=1,2,\ldots,N\}\in \mathbb{R}^{M\times N}$, which can also be expressed as $\boldsymbol{\hat{Z}}=\{\hat{z}_1,\ldots,\hat{z}_i,\ldots,\hat{z}_L|i=1,2,\ldots,L\}\in \mathbb{R}^{L\times 1}$.

\subsubsection{Anomaly Detection}
As shown in Fig. \ref{fig:4}, to obtain the anomaly detection results, a complete forward propagation and backpropagation are necessary. Substituting the test HSI $\boldsymbol{H}$ and the trained student network $f_s (\cdot;\theta_s)$ into Equation \ref{e:6}, we have

\begin{equation}
	\label{e:12}
	\boldsymbol{\hat{H}}_1,\boldsymbol{\hat{H}}_2,\boldsymbol{\hat{H}}_3=f_s(\boldsymbol{H};\theta_s).
\end{equation}

To prevent the backward transmission of erroneous results, we mask a portion of the neurons by incorporating the mask matrix $\boldsymbol{\hat{Z}}$ calculated in the previous section to the loss function. Before implementing this, we need to set the smaller element of $d$ to zero by

\begin{equation}
	\label{e:13}
	\hat{z}_i=
	\begin{cases} 
		\hat{z}_i,  & \hat{z}_i > \hat{z}_m \\
		0, & \hat{z}_i \leq \hat{z}_m
	\end{cases},
\end{equation}
where $\hat{z}_m$ denotes the median of $\boldsymbol{\hat{Z}}$. The partially masked loss function is then computed as

\begin{equation}
	\label{e:14}
	\mathcal{L}_s = {\rm sum}\left(\sum_{k=1,2,3} \left(\boldsymbol{H}-\boldsymbol{\hat{H}}_k \right)^2\odot \boldsymbol{\hat{Z}}\right).
\end{equation}

According to Equation \ref{e:2}, the bias derivative of the test HSI with respect to $\mathcal{L}_s$ is calculated as

\begin{equation}
	\label{e:15}
	\boldsymbol{G} = \frac{\partial \mathcal{L}_s}{\partial \boldsymbol{H}}.
\end{equation}

Finally, the detection results can be calculated according to

\begin{equation}
	\label{e:16}
	\boldsymbol{S} = {\rm max} (|\boldsymbol{G}|),
\end{equation}
where the computation of the maximum value is performed in the band dimension. The details of our method are outlined in Algorithm \ref{a:1}.

\newcommand\mycommfont[1]{\footnotesize\ttfamily\textcolor{blue}{#1}}
\SetCommentSty{mycommfont}

\SetKwInput{KwInput}{Input}                
\SetKwInput{KwOutput}{Output}              
\begin{algorithm}
	\label{a:1}
	\caption{Knowledge Distillation and STAD Detection}
	\KwIn{Initialized teacher network $f_t(\cdot;\theta_t)$ and student network $f_s(\cdot;\theta_s)$, training set $\boldsymbol{D}$, test HSI $\boldsymbol{H}$}
	\KwOut{Test result $\boldsymbol{S}$}
	\tcc{Training the teacher network}
	\For{$epoch=1,2,\ldots,eopchs$}{
		$\boldsymbol{\hat{D}}_{t1},\boldsymbol{\hat{D}}_{t2},\boldsymbol{\hat{D}}_{t3}=f_t(\boldsymbol{D};\theta_t)$ \tcp*{Equation \ref{e:4}}
	
		$\mathcal{L}_t=\sum_{k=1,2,3} \left \| \boldsymbol{\hat{D}}_{tk} - \boldsymbol{D} \right \|_2$ \tcp*{Equation \ref{e:5}}
	
		$grad_t = \frac{\partial \mathcal{L}_t}{\partial \theta_t}$
	
		Update $\theta_t$ based on $grad_t$
	} 
	\tcc{Training the student network}
	\For{$epoch=1,2,\ldots,eopchs$}{
		$\boldsymbol{\hat{D}}_{s1},\boldsymbol{\hat{D}}_{s2},\boldsymbol{\hat{D}}_{s3}=f_s(\boldsymbol{D};\theta_s)$ \tcp*{Equation \ref{e:6}}
		
		$\mathcal{L}_{kd}=\sum_{k=1,2,3} \left \| \boldsymbol{\hat{D}}_{tk} - \boldsymbol{\hat{D}}_{sk} \right \|_2$ \tcp*{Equation \ref{e:7}}
		
		$grad_s = \frac{\partial \mathcal{L}_kd}{\partial \theta_s}$
		
		Update $\theta_s$ based on $grad_s$
	} 
	\tcc{Detection of anomalies}
	$\boldsymbol{Z}\leftarrow$Mahalanobis distance of $\boldsymbol{H}$ \tcp*{Equation \ref{e:8},\ref{e:9}}
	
	$\boldsymbol{\hat{Z}}\leftarrow$Bilateral filtering on $\boldsymbol{Z}$ \tcp*{Equation \ref{e:10},\ref{e:11}}
	
	Elements of $\boldsymbol{\hat{Z}}$ less than median $\rightarrow 0$ \tcp*{Equation \ref{e:13}}
	
	$\boldsymbol{\hat{H}}_1, \boldsymbol{\hat{H}}_2, \boldsymbol{\hat{H}}_3 = f_s(\boldsymbol{H};\theta_s)$ \tcp*{Equation \ref{e:12}}
	
	Calculate the masked loss $\mathcal{L}_s$ \tcp*{Equation \ref{e:14}}
	
	$grad = \frac{\partial \mathcal{L}_s}{\partial \boldsymbol{H}}$ \tcp*{Equation \ref{e:15}}
	
	$\boldsymbol{S}\leftarrow$ Maximize in each band of $grad$ \tcp*{Equation \ref{e:16}}
	
	\Return{$\boldsymbol{S}$}
\end{algorithm}

\subsection{HAD with Human Visual Perception}
The network structure, visualization method, and activation method employed in our proposed method are relatively basic, and it's evident that more advanced and effective methods exist. Hence, in this section, we explore alternative ways to apply human visual perception to HAD. These explorations may provide inspiration for future research.

\subsubsection{Methods Based on Estimation}
\begin{figure}[t]
	\centering
	\includegraphics[width=0.5\textwidth]{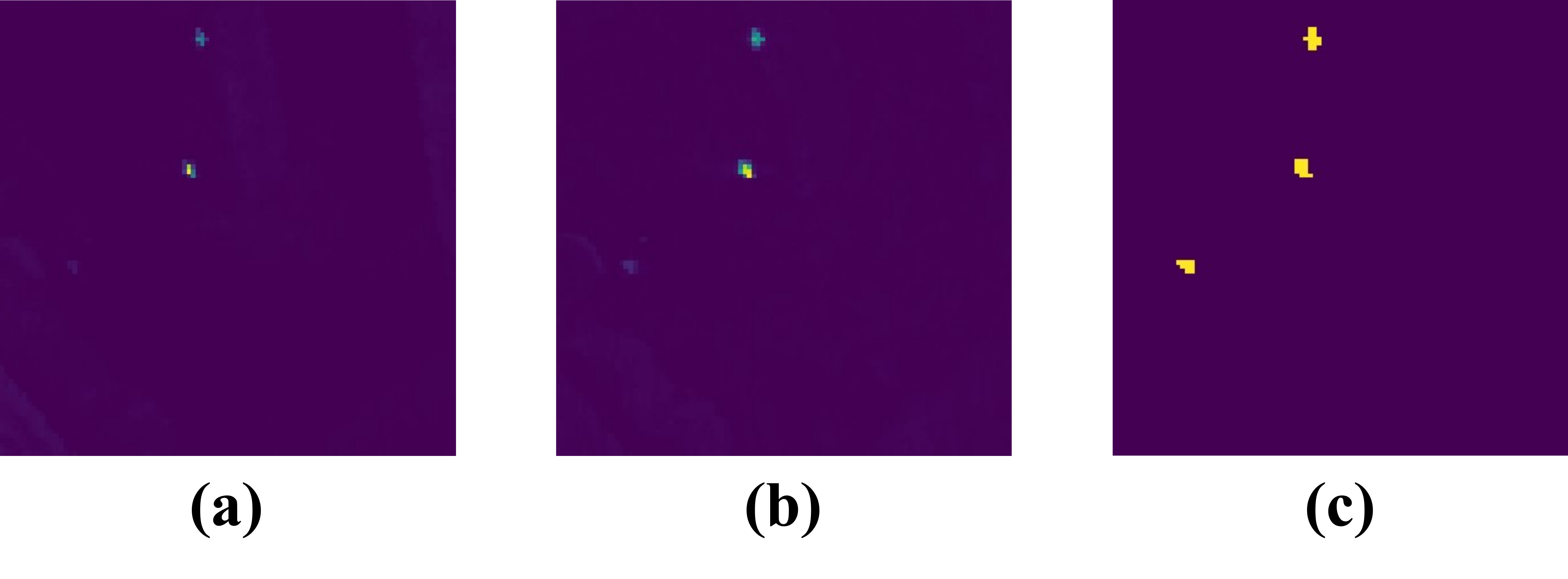}
	\caption{One of the results of testing on the HAD100 dataset. (a) Result of estimation-based method. (b) Results of the saliency map based method, more anomalies are detected compared to estimation-based method. (c) Anomalies map.}
	\label{fig:5}
\end{figure}

Different network structures may lead to different feature spaces. We take the estimation-based HAD method as an example. Although the estimation-based methods use features from the middle layer of the network, the primary reason for this is the low-dimensional property of the features. However, some information or anomalous representations are lost in the process of feature extraction to density estimation. This usually results in a lower detection rate of anomalies, which can be improved by fully utilizing the feature space, as shown in Fig. \ref{fig:5}. To demonstrate, we construct a simple reconstruction network and density estimation network for HAD, both of which are fully connected networks. Traditional estimation-based methods output anomaly score maps composed of negative log-likelihood estimates, also known as probability density energy maps. The loss function is a weighted sum of the reconstruction error, probability density and penalty term. Among them, it is the probability density that is directly related to the detection result. Therefore, for testing, we multiply and average the mask matrix only with the probability density matrix as the loss for backpropagation. As shown in Fig. \ref{fig:5}, the saliency map detects more anomalous pixels. Regarding other network structures, it is necessary to determine the way of backpropagation to obtain the saliency map based on the calculation of the anomaly score map, which is one of the directions of the subsequent research. 

\subsubsection{Visual Perception Feature Space}
As mentioned above, saliency maps provide a simple way to visualize feature maps. We introduce in Section \ref{2a} several more recent methods for visualizing feature maps. These methods are capable of extracting maps from the feature space that align more with human visual perception and facilitate better localization of anomalies. Additionally, the selection of the mask matrix is an important research direction. In this paper, we use the mask matrix with the aim of providing small target localization, which is only a special consideration for activation methods. Building on this, a refined masking matrix enables more precise masking of neurons that negatively impact detection. In summary, to further explore HAD consistent with human visual perception, different network structures, visualization methods, and activation methods can be considered.

\section{Experiments}
In this section, extensive experiments are conducted to evaluate and validate the effectiveness and generalization of the proposed STAD method. It is compared qualitatively and quantitatively with the most advanced and typical HAD methods.

\subsection{Experimental Settings}
\subsubsection{Datasets}
We validate our method on the HAD100 [aetnet] dataset. This dateset contains multiple HSIs captured by the AVIRIS sensor. It provides two training sets and one test set, where the training set we use is the one downloaded from AVIRIS-NG. The training set contains 260 HSIs without anomalies and the test set contains 100 HSIs with anomalies and their labels. All HSIs in the training set have a size of 81$\times$81 with 425 bands. The size of HSIs in the test set varies from 64$\times$64 to 101$\times$101 with 425 bands. Note that we just used the official 94 original HSIs with anomalies and did not crop them to 100 HSIs shaped as 64$\times$64. In addition, since the original HSIs contain a large number of noisy bands, we only use the first 50 of them.

\subsubsection{Comparison Methods}
Seven frequently cited and state-of-the-art AD methods are compared, including three deep learning-based methods, Auto-AD \cite{autoad}, DFAE \cite{dfae}, AETNet \cite{aetnet}, and five traditional-based methods, RX \cite{rx}, CRD \cite{crd}, LRASR \cite{lrasr}, AED \cite{aed}, PTA \cite{pta}. The input and output window sizes of CRD are set to 5 and 29, respectively. The beta and lamda of LRASR are set to 0.1 and 1, respectively. The number of downscaled bands and range standard deviation of AED are set to 30 and 0.3, respectively. All other parameters were set to remain the same as in the original paper.

\subsubsection{Evaluation Metrics}
In this article, we use the area under the curve (AUC) \cite{auc} of the receiver operating characteristic (ROC) \cite{roc} to compare the detection performance of different AD methods on the HAD100 dataset. AUC is the most widely used evaluation metric in HAD tasks and can directly provide quantitative detection accuracy. In addition, since the HAD100 test set contains close to 100 HSIs, there is not enough space to use Box-Whisker Plot \cite{box} to represent the separability of backgrounds and anomalies. Therefore, we apply the background suppressibility to quantitatively assess anomaly separability.

3D-ROC curves can comprehensively quantify the accuracy, false alarm rate and anomaly separability of test results. Given the segmentation threshold $\tau$, pixels with scores greater than $\tau$ in the detection results are classified as positive samples, then the detection probability $P_d$ and the false detection probability $P_f$ are computed according to

\begin{equation}
	P_d = \frac{TP}{TP+FN},
\end{equation}

\begin{equation}
	P_f = \frac{FP}{TN+FP},
\end{equation}
where $TP$ denotes the number of true-positive samples, $FP$ denotes the number of false-positive samples, $FN$ denotes the number of false-negative samples, $TN$ denotes the number of true-negative samples. Each $\tau$ has a corresponding pair ($P_d, P_f$), represented on the 3D-ROC curve as ($P_d, P_f, \tau$). Therefore, three 2D-ROC curves can be obtained from the 3D-ROC curves as ($P_d, P_f$), ($P_d, \tau$), ($P_f, \tau$). With $p=30000$ points sample from the curve at equal interval, the AUC value of ($P_d, P_f$) indicates the detection accuracy, calculated as

\begin{equation}
	AUC_{(D,F)} = \frac{1}{2}\sum_{i=1}^{p-1}(P_d^{i+1} + P_d^{i})(P_f^{i+1} - P_f^{i}).
\end{equation}

Higher values of $AUC_{(D,F)}$ suggest superior performance in detection. The AUC value of ($P_f, \tau$) indicates false alarm rate, calculated as

\begin{equation}
	AUC_{(F,\tau)} = \frac{1}{2}\sum_{i=1}^{p-1}(P_f^{i} + P_f^{p-i})(\tau^{i+1} - \tau^{i}).
\end{equation}

Lower values of $AUC_{(F,\tau)}$ indicate superior background suppression performance. The background suppressibility value $AUC_{BS}$ is calculated as 

\begin{equation}
	AUC_{BS} = AUC_{(D,F)} - AUC_{(F,\tau)}.
\end{equation}

As with detection accuracy, this value is positively correlated with detection performance.

\subsubsection{Implementation Details of STAD}
Our teacher network takes Transformer as backbone which has 3 layers, 1000 hidden dimensions and 2 heads in the self-attention layer. The teacher network is trained using the Adam \cite{adam} optimizer with a learning rate of $10^{-4}$, epochs set to 50 and batch size set to 16. The student network consists of a convolutional layer, a deconvolutional layer and a fully-connected layer with 100 hidden dimensions. The student network is trained using the Adam optimizer with a learning rate of $10^{-4}$, epochs set to 250 and batch size set to 16. All models are trained using an exponential moving \cite{ema} average algorithm with a decay factor of 0.9. For STF, $r$ is set to 1, $\sigma_s$ is set to 1, and $\sigma_c$ is set to 80. Note that the setting of $\sigma_c$ is based on the input image in 8-bit UINT format.

\subsection{Effectiveness Assessment}
The comparison results of the detection performance are listed in Table \ref{t:1}. To broadly assess the effectiveness of HAD methods, we categorize detection results into four groups: top-performing (underlined), second-best (blue), third best (red), and failures (gray). The latter indicate a failure to adequately identify enough anomalous targets or suppress the background. As shown in Table \ref{t:1}, our method achieves the highest average $AUC_{(D,F)}$ of 0.9899, followed by AETNet with 0.9877 and LRX with 0.9730. Taking advantage of the fact that STAD retains more anomalous representations, our method achieves high accuracy on most datasets, even higher than AETNet, which was designed specifically for the HAD100 dataset. Other methods, such as DFAE, may only attain high detection accuracy on certain HSIs, while on other HSIs, detection accuracy is low or anomalous targets may remain undetected. This is attributed to the limited generalization of these methods, as optimal accuracy cannot be attained by employing identical sets of parameters across all HSIs. To illustrate the performance of the compared methods more intuitively, we statistically present their experimental results in Table \ref{t:2}. In terms of $AUC_{(D,F)}$, our method achieves the highest percentage across all three performance tops, which is consistent with our previous description. Furthermore, since STAD uses a feature space with firm perceptual representation, our method exhibits only two failures in detection out of 94 HSIs, with the lowest $AUC_{(D,F)}$ recorded at 0.8501. This also indicates that the feature space has stronger robustness.

In terms of background suppression performance, STAD also achieves the highest $AUC_{BS}$ of 0.9836, the second highest is AETNet at 0.9538, and the third highest is LRX at 0.9436, as shown in Table \ref{t:2}. Specifically, the $AUC_{(D,F)}$ and $AUC_{BS}$ results for STAD exhibit similarity, with a marginal difference of only 0.0063. This observation suggests that our method maintains a low false alarm rate. DFAE demonstrates a significantly reduced false alarm rate, possibly attributed to the substantial enhancement in contrast achieved by its exponential fusion component. However, this heightened contrast also results in the suppression of certain anomalous targets, consequently diminishing overall detection accuracy. Although AETNet has the second-highest $AUC_{BS}$, it performs poorly in background suppression and has a false alarm rate 0.0298 lower than STAD. Similarly, we present a more intuitive statistical presentation in Table \ref{t:2}. As in our analysis, DFAE has better background suppressibility, but the results are unstable, with failures on 11 HSIs. The results of AETNet are stable, but the overall false alarm rate is high. While our STAD method exhibits higher stability and lower false alarm rate due to the strong robustness of its feature space and the masking of low confidence activations by STF.

\begin{figure*}[t]
	\centering
	\includegraphics[width=1.0\textwidth]{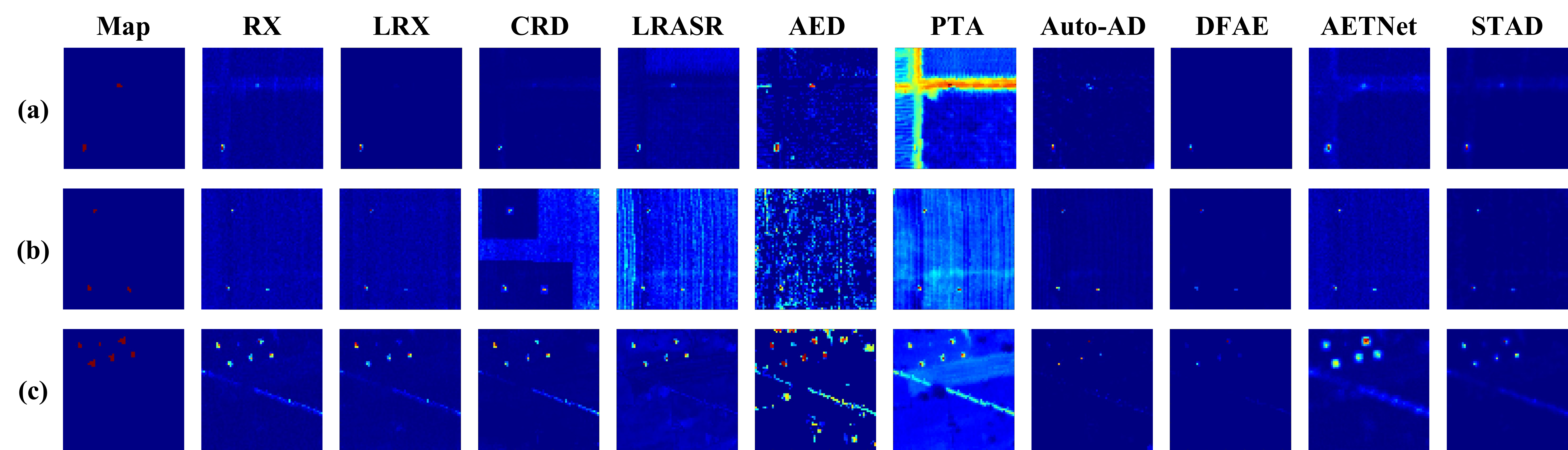}
	\caption{Anomaly maps, and detection maps of the compared methods for the HSIs in HAD100 dataset, with indexes of (a) 59, (b) 77, (c) 92.}
	\label{fig:6}
\end{figure*}

Furthermore, we chose three HSIs from the test set, collected in different time periods and scenarios, to visualize their detection results using different methods, as shown in Figure \ref{fig:6}. Our method retains anomalous targets more completely compared to traditional methods, and no targets go undetected. Compared to other deep learning methods besides DFAE, our method exhibits a lower bounded response and no significant background. In contrast, the results of DFAE and Auto-AD exhibited comparable characteristics to those of traditional methods, with some targets remaining undetected. Our proposed approach emphasizes the advantages of achieving comprehensive and accurate HAD. This observation indicates promising potential for the development of HAD techniques based on feature space. Furthermore, the visualization results, crafted to be more in harmony with human vision, confer practical advantages to our method in real-world applications. This alignment with human vision enhances the interpretability and usability of our approach, contributing to its efficacy in remote sensing scenarios.

\begin{table*}[]
	\centering
	\caption{Evaluation of the Detection Accuracy and Background Suppressibility of Different Methods on the HAD100 Dataset. \underline{Underlined cells} Indicate the Corresponding \underline{Optimal} Performance, and {\color{blue} Blue} and {\color{red} Red} Colors Indicate the {\color{blue} Second} and {\color{red} Third} Best Performance, Respectively. Cells with a Light Gray Background Indicate that the Corresponding Method Fails on This Image, Judged by a Detection Accuracy of Less Than 0.9 or a Background Suppressibility of Less Than 0.8.}
	\tiny
	\tabcolsep=0.15cm
	\begin{tabular}{c|cccccccccc|cccccccccc}
		\toprule[0.5mm]
		~ & \multicolumn{10}{c|}{$AUC_{(D,F)}$} & \multicolumn{10}{c}{$AUC_{BS}$} \\ \midrule
		Index & RX & LRX & CRD & LRASR & AED & PTA & Auto-AD & DFAE & AETNet & \textbf{STAD} & RX & LRX & CRD & LRASR & AED & PTA & Auto-AD & DFAE & AETNet & \textbf{STAD} \\ \midrule
		1 & \color{red}0.9984 & \underline{0.9996} & \cellcolor{gray!40}0.7419 & 0.9635 & 0.9041 & 0.9756 & \cellcolor{gray!40}0.8516 & 0.9911 & 0.9892 & \color{blue}0.9988 & 0.9863 & \color{blue}0.9969 & \cellcolor{gray!40}0.7377 & 0.9369 & 0.8654 & 0.8246 & 0.8504 & \color{red}0.9911 & 0.9785 & \underline{0.9972} \\
		2 & \underline{1.0000} & 0.9995 & 0.9991 & \color{blue}0.9999 & 0.9951 & \underline{1.0000} & 0.9507 & 0.9983 & 0.9962 & \color{red}0.9998 & 0.9622 & 0.9661 & \color{red}0.9872 & 0.9569 & 0.9728 & 0.8209 & 0.9496 & \underline{0.9980} & 0.9667 & \color{blue}0.9953 \\
		3 & \color{blue}0.9880 & \underline{0.9926} & \cellcolor{gray!40}0.5038 & \cellcolor{gray!40}0.6319 & 0.9714 & 0.9150 & 0.9690 & 0.9341 & 0.9800 & \color{red}0.9859 & \color{red}0.9752 & \underline{0.9870} & \cellcolor{gray!40}0.5038 & \cellcolor{gray!40}0.5585 & 0.9601 & 0.8370 & 0.9684 & 0.9341 & 0.9637 & \color{blue}0.9833 \\
		4 & \color{blue}0.9997 & 0.9982 & 0.9708 & 0.9951 & 0.9961 & 0.9958 & 0.9744 & \underline{0.9998} & 0.9981 & \color{red}0.9996 & 0.9745 & 0.9755 & 0.9589 & 0.9370 & 0.9473 & 0.8486 & 0.9723 & \underline{0.9998} & \color{red}0.9839 & \color{blue}0.9969 \\
		5 & 0.9560 & 0.9683 & 0.9884 & 0.9951 & \underline{0.9988} & 0.9245 & \color{red}0.9979 & \color{blue}0.9986 & 0.9920 & 0.9946 & 0.8825 & 0.9060 & 0.9765 & 0.9387 & 0.9538 & \cellcolor{gray!40}0.5922 & \color{blue}0.9922 & \underline{0.9984} & 0.9268 & \color{red}0.9889 \\
		6 & 0.9912 & 0.9578 & \cellcolor{gray!40}0.8616 & 0.9894 & 0.9697 & 0.9572 & 0.9758 & \color{blue}0.9973 & \color{red}0.9950 & \underline{0.9976} & 0.9744 & 0.9421 & 0.8520 & 0.9619 & 0.9560 & \cellcolor{gray!40}0.7990 & 0.9748 & \underline{0.9973} & \color{red}0.9751 & \color{blue}0.9942 \\
		7 & 0.9960 & 0.9960 & 0.9706 & 0.9243 & 0.9894 & \cellcolor{gray!40}0.7323 & 0.9528 & \underline{0.9993} & \color{red}0.9988 & \color{blue}0.9990 & 0.9470 & 0.9584 & 0.9511 & 0.8214 & 0.9547 & \cellcolor{gray!40}0.4352 & 0.9516 & \underline{0.9992} & \color{red}0.9754 & \color{blue}0.9942 \\
		8 & \underline{1.0000} & \underline{1.0000} & 0.9975 & 0.9990 & 0.9920 & 0.9848 & 0.9936 & \color{blue}0.9998 & \color{red}0.9997 & \underline{1.0000} & 0.9574 
		& 0.9881 & 0.9877 & 0.8891 & 0.9250 & \cellcolor{gray!40}0.7766 & \color{red}0.9906 & \underline{0.9997} & 0.9638 & \color{blue}0.9965 \\
		9 & 0.9979 & \color{blue}0.9996 & 0.9849 & 0.9874 & 0.9740 & \underline{1.0000} & 0.9549 & \cellcolor{gray!40}0.5485 & \color{red}0.9984 & \underline{1.0000} & 
		0.9092 & \color{blue}0.9606 & \color{red}0.9591 & 0.8898 & 0.9207 & 0.8205 & 0.9475 & \cellcolor{gray!40}0.5485 & 0.9554 & \underline{0.9930} \\
		10 & 0.9770 & 0.9817 & 0.9199 & 0.9798 & 0.9756 & \color{red}0.9988 & 0.9917 & \color{red}0.9988 & \color{blue}0.9992 & \underline{0.9998} & 0.9162 & 0.8842 & 0.8692 & 0.8317 & 0.8664 & 0.8733 & \color{red}0.9891 & \underline{0.9985} & 0.9527 & \color{blue}0.9958 \\
		11 & \color{blue}0.9997 & 0.9982 & \cellcolor{gray!40}0.8095 & \cellcolor{gray!40}0.8921 & \color{red}0.9996 & \cellcolor{gray!40}0.8992 & 0.9911 & 0.9990 & 0.9988 & \underline{0.9998} & 0.9838 & \color{red}0.9933 & 0.8040 & 0.8589 & 0.9838 & 0.8265 & 0.9906 & \underline{0.9990} & 0.9847 & \color{blue}0.9973 \\        
		12 & \color{blue}0.9996 & 0.9961 & \cellcolor{gray!40}0.5581 & \color{red}0.9995 & 0.9638 & \cellcolor{gray!40}0.7953 & 0.9886 & 0.9977 & 0.9978 & \underline{0.9999} & 0.9718 & \color{red}0.9896 & \cellcolor{gray!40}0.5485 & 0.9503 & 0.8996 & \cellcolor{gray!40}0.7150 & 0.9880 & \underline{0.9977} & 0.9852 & \color{blue}0.9968 \\
		13 & \color{red}0.9666 & 0.9372 & \cellcolor{gray!40}0.7573 & \cellcolor{gray!40}0.7493 & \cellcolor{gray!40}0.8889 & \cellcolor{gray!40}0.8485 & 0.9335 & 0.9281 & \color{blue}0.9738 & \underline{0.9853} & \color{red}0.9464 & 0.9367 & \cellcolor{gray!40}0.7489 & \cellcolor{gray!40}0.6963 & 0.8736 & \cellcolor{gray!40}0.6817 & 0.9326 & 0.9281 & \color{blue}0.9560 & \underline{0.9820} \\
		14 & \underline{1.0000} & \color{blue}0.9999 & 0.9916 & \underline{1.0000} & 0.9990 & \cellcolor{gray!40}0.8999 & \color{red}0.9992 & \underline{1.0000} & 0.9990 & \color{blue}0.9999 & 0.9667 & 0.9838 & 0.9803 & 0.9479 & 0.9828 & \cellcolor{gray!40}0.6926 & \color{blue}0.9982 & \underline{1.0000} & 0.9814 & \color{red}0.9976 \\
		15 & 0.9886 & 0.9913 & 0.9077 & 0.9990 & 0.9948 & \cellcolor{gray!40}0.6303 & 0.9989 & \underline{1.0000} & \color{blue}0.9998 & \color{red}0.9991 & 0.9431 & 0.8746 & 0.8708 & 0.9168 & 0.9683 & \cellcolor{gray!40}0.2598 & \color{red}0.9942 & \underline{1.0000} & 0.9726 & \color{blue}0.9955 \\
		16 & \underline{0.9999} & 0.9540 & 0.9234 & \color{red}0.9994 & \cellcolor{gray!40}0.8149 & \cellcolor{gray!40}0.8519 & 0.9906 & 0.9989 & 0.9974 & \color{blue}0.9995 & 0.9844 & 0.9357 & 0.9205 & 0.9823 & \cellcolor{gray!40}0.7691 & \cellcolor{gray!40}0.7384 & \color{red}0.9903 & \underline{0.9989} & 0.9849 & \color{blue}0.9975 \\
		17 & \color{blue}0.9998 & 0.9994 & 0.9877 & 0.9923 & \color{red}0.9997 & 0.9913 & 0.9670 & 0.9910 & 0.9993 & \underline{1.0000} & 0.9109 & 0.8785 & 0.9508 & 0.9017 & 0.9534 & \cellcolor{gray!40}0.6472 & \color{red}0.9612 & \color{blue}0.9905 & 0.9322 & \underline{0.9929} \\
		18 & 0.9930 & 0.9555 & 0.9795 & 0.9798 & \underline{0.9996} & 0.9573 & \cellcolor{gray!40}0.7269 & 0.9833 & \color{blue}0.9979 & \color{red}0.9935 & 0.8860 & 0.8145 & 0.9338 & 0.8558 & \color{red}0.9567 & \cellcolor{gray!40}0.6027 & \cellcolor{gray!40}0.7009 & \underline{0.9831} & 0.9044 & \color{blue}0.9802 \\        
		19 & \color{blue}0.9994 & \color{red}0.9993 & 0.9757 & 0.9699 & 0.9756 & 0.9986 & 0.9881 & 0.9553 & 0.9956 & \underline{0.9997} & 0.9752 & \color{blue}0.9934 & 
		0.9672 & 0.9221 & 0.9651 & 0.8994 & \color{red}0.9876 & 0.9553 & 0.9846 & \underline{0.9953} \\
		20 & 0.9581 & 0.9626 & 0.9064 & 0.9906 & 0.9709 & 0.9044 & 0.9425 & \underline{0.9977} & \color{red}0.9952 & \color{blue}0.9956 & 0.9332 & 0.9584 & 0.9016 & 0.9550 & 0.9189 & \cellcolor{gray!40}0.7760 & 0.9402 & \underline{0.9977} & \color{red}0.9781 & \color{blue}0.9907 \\
		21 & \underline{1.0000} & \underline{1.0000} & \color{blue}0.9999 & \underline{1.0000} & \color{red}0.9998 & 0.9982 & 0.9901 & \underline{1.0000} & 0.9982 & \color{blue}0.9999 & 0.9812 & \color{red}0.9959 & 0.9950 & 0.9518 & 0.9694 & 0.8706 & 0.9891 & \underline{1.0000} & 0.9807 & \color{blue}0.9972 \\
		22 & \color{red}0.9985 & 0.9971 & 0.9780 & 0.9962 & 0.9971 & 0.9803 & 0.9611 & \color{red}0.9985 & \color{blue}0.9986 & \underline{0.9993} & 0.9729 & \color{blue}0.9952 & 0.9745 & 0.9749 & 0.9575 & \cellcolor{gray!40}0.7674 & 0.9579 & \underline{0.9984} & 0.9749 & \color{red}0.9945 \\
		23 & 0.9950 & \color{blue}0.9982 & \cellcolor{gray!40}0.8691 & 0.9878 & 0.9941 & 0.9740 & 0.9500 & \color{red}0.9973 & 0.9944 & \underline{0.9989} & 0.9785 & \color{blue}0.9965 & 0.8558 & 0.9519 & 0.9756 & \cellcolor{gray!40}0.7796 & 0.9479 & \underline{0.9973} & 0.9737 & \color{red}0.9963 \\
		24 & 0.9983 & 0.9946 & \cellcolor{gray!40}0.8398 & 0.9810 & 0.9895 & \color{red}0.9987 & 0.9740 & \color{blue}0.9988 & 0.9984 & \underline{0.9992} & 0.9679 & 0.9690 & 0.8301 & 0.9460 & 0.9656 & 0.9464 & 0.9733 & \underline{0.9988} & \color{red}0.9799 & \color{blue}0.9945 \\
		25 & 0.9984 & \underline{1.0000} & \cellcolor{gray!40}0.5442 & 0.9938 & \color{blue}0.9993 & 0.9945 & \color{red}0.9989 & 0.9969 & 0.9948 & 0.9981 & 0.9916 & \underline{0.9991} & \cellcolor{gray!40}0.5442 & 0.9732 & 0.9951 & 0.9236 & \color{blue}0.9987 & \color{red}0.9969 & 0.9825 & 0.9957 \\
		26 & 0.9993 & \color{red}0.9995 & 0.9553 & \color{blue}0.9998 & 0.9981 & \cellcolor{gray!40}0.5864 & 0.9622 & 0.9992 & 0.9995 & \underline{0.9999} & 0.9557 & \color{red}0.9891 & 0.9231 & 0.9616 & 0.9755 & \cellcolor{gray!40}0.1666 & 0.9593 & \underline{0.9992} & 0.9768 & \color{blue}0.9962 \\
		27 & \underline{0.9998} & \color{red}0.9993 & \cellcolor{gray!40}0.6001 & 0.9760 & 0.9960 & 0.9976 & 0.9946 & 0.9992 & 0.9976 & \color{blue}0.9996 & 0.9872 & \color{blue}0.9979 & \cellcolor{gray!40}0.6001 & 0.9408 & 0.9856 & 0.9288 & 0.9942 & \underline{0.9992} & 0.9885 & \color{red}0.9974 \\
		28 & 0.9840 & 0.9716 & 0.9546 & 0.9697 & \underline{0.9990} & 0.9593 & 0.9692 & \color{red}0.9978 & \color{blue}0.9990 & 0.9916 & 0.8928 & 0.9267 & 0.9317 & 0.8513 & \color{red}0.9676 & \cellcolor{gray!40}0.7812 & 0.9655 & \underline{0.9978} & 0.9562 & \color{blue}0.9796 \\
		29 & 0.9910 & 0.9920 & \cellcolor{gray!40}0.7713 & 0.9512 & 0.9851 & \cellcolor{gray!40}0.5447 & 0.9624 & \color{red}0.9924 & \color{blue}0.9972 & \underline{0.9988} & 0.9495 & \color{red}0.9791 & \cellcolor{gray!40}0.7470 & 0.8200 & 0.9695 & \cellcolor{gray!40}0.2487 & 0.9603 & \color{blue}0.9923 & 0.9687 & \underline{0.9951} \\
		30 & \color{blue}0.9976 & 0.9887 & 0.9425 & 0.9311 & 0.9793 & \color{red}0.9959 & \cellcolor{gray!40}0.8683 & \cellcolor{gray!40}0.8024 & 0.9959 & \underline{0.9996} & 0.9701 & \color{red}0.9732 & 0.9239 & 0.8096 & 0.9073 & 0.8824 & 0.8604 & 0.8024 & \color{blue}0.9760 & \underline{0.9931} \\
		31 & 0.9807 & 0.9896 & \cellcolor{gray!40}0.8472 & 0.9647 & 0.9189 & 0.9533 & \cellcolor{gray!40}0.8669 & \underline{0.9973} & \color{red}0.9930 & \color{blue}0.9965 & 0.9607 & \color{red}0.9774 & 0.8392 & 0.9365 & 0.8728 & \cellcolor{gray!40}0.7804 & 0.8651 & \underline{0.9973} & 0.9765 & \color{blue}0.9940 \\        
		32 & \underline{0.9999} & \underline{0.9999} & 0.9408 & \color{blue}0.9997 & \cellcolor{gray!40}0.5467 & \color{red}0.9990 & 0.9812 & 0.9585 & 0.9943 & 0.9983 & \color{blue}0.9839 & \underline{0.9959} & 0.9374 & 0.9764 & \cellcolor{gray!40}0.4627 & 0.8912 & 0.9797 & 0.9585 & \color{red}0.9817 & \underline{0.9959} \\   
		33 & 0.9758 & \underline{0.9856} & \cellcolor{gray!40}0.5157 & \cellcolor{gray!40}0.8522 & 0.9157 & 0.9477 & \color{red}0.9824 & \cellcolor{gray!40}0.8143 & \color{blue}0.9852 & 0.9744 & 0.9641 & \color{blue}0.9757 & \cellcolor{gray!40}0.5157 & 0.8149 & 0.9058 & 0.8704 & \underline{0.9784} & 0.8143 & \color{red}0.9713 
		& 0.9702 \\
		34 & \color{blue}0.9988 & 0.9878 & 0.9905 & \cellcolor{gray!40}0.7988 & \cellcolor{gray!40}0.7932 & \color{red}0.9956 & 0.9142 & \cellcolor{gray!40}0.7394 & 0.9920 & \underline{0.9990} & 0.9159 & 0.9149 & \color{blue}0.9717 & \cellcolor{gray!40}0.6365 & \cellcolor{gray!40}0.7808 & \cellcolor{gray!40}0.7919 & 0.9103 & \cellcolor{gray!40}0.7394 & \color{red}0.9502 & \underline{0.9921} \\
		35 & 0.9620 & 0.9484 & \color{blue}0.9855 & 0.9363 & \underline{0.9956} & 0.9202 & \underline{0.9956} & \color{red}0.9789 & 0.9764 & 0.9597 & 0.9500 & 0.9463 & 
		\color{blue}0.9825 & 0.8908 & 0.9624 & \cellcolor{gray!40}0.4893 & \underline{0.9852} & \color{red}0.9789 & 0.9625 & 0.9586 \\
		36 & 0.9982 & 0.9984 & 0.9983 & 0.9177 & 0.9047 & 0.9960 & 0.9833 & \underline{0.9999} & \color{red}0.9989 & \color{blue}0.9997 & 0.9286 & 0.9510 & \color{red}0.9831 & 0.8372 & 0.8366 & 0.8169 & 0.9804 & \underline{0.9999} & 0.9829 & \color{blue}0.9912 \\
		37 & \color{blue}0.9998 & 0.9847 & 0.9672 & \color{blue}0.9998 & 0.9983 & \cellcolor{gray!40}0.7749 & 0.9931 & 0.9977 & \color{red}0.9998 & \underline{0.9999} & 0.9750 & 0.8748 & 0.9381 & 0.9330 & 0.9840 & \cellcolor{gray!40}0.4279 & \color{red}0.9909 & \underline{0.9974} & 0.9825 & \color{blue}0.9971 \\
		38 & \underline{0.9983} & 0.9975 & \cellcolor{gray!40}0.5284 & 0.9135 & 0.9975 & \color{blue}0.9980 & 0.9668 & 0.9840 & 0.9936 & \color{red}0.9979 & 0.9789 & \color{blue}0.9895 & \cellcolor{gray!40}0.5284 & 0.8795 & 0.9805 & 0.9157 & 0.9656 & \color{red}0.9840 & 0.9810 & \underline{0.9948} \\
		39 & \color{red}0.9802 & 0.9671 & \cellcolor{gray!40}0.7888 & \cellcolor{gray!40}0.7275 & \cellcolor{gray!40}0.8531 & 0.9287 & 0.9569 & 0.9440 & \underline{0.9943} & \color{blue}0.9896 & 0.9489 & 0.9365 & \cellcolor{gray!40}0.7771 & \cellcolor{gray!40}0.5500 & 0.8395 & \cellcolor{gray!40}0.7704 & \color{red}0.9551 & 0.9440 & \color{blue}0.9657 & \underline{0.9831} \\
		40 & \color{red}0.9959 & \color{blue}0.9978 & 0.9571 & 0.9220 & 0.9202 & 0.9634 & 0.9173 & \cellcolor{gray!40}0.7446 & 0.9910 & \underline{0.9990} & 0.9562 & \color{blue}0.9723 & 0.9479 & 0.8214 & 0.8914 & \cellcolor{gray!40}0.7881 & 0.9139 & \cellcolor{gray!40}0.7446 & \color{red}0.9621 & \underline{0.9915} \\        
		41 & 0.9914 & 0.9516 & \cellcolor{gray!40}0.8977 & 0.9758 & 0.9939 & \cellcolor{gray!40}0.4934 & 0.9553 & \color{red}0.9961 & \color{blue}0.9977 & \underline{0.9997} & 0.9466 & 0.9056 & 0.8394 & 0.8651 & \color{red}0.9641 & \cellcolor{gray!40}0.0594 & 0.9398 & \underline{0.9960} & 0.9550 & \color{blue}0.9921 \\        
		42 & \cellcolor{gray!40}0.8730 & 0.9049 & \cellcolor{gray!40}0.8074 & 0.9255 & \color{red}0.9839 & \cellcolor{gray!40}0.6914 & 0.9728 & \underline{0.9972} & 0.9811 & \color{blue}0.9872 & \cellcolor{gray!40}0.7763 & 0.8326 & \cellcolor{gray!40}0.7673 & 0.8073 & 0.9232 & \cellcolor{gray!40}0.3929 & \color{red}0.9675 & \underline{0.9971} & 0.9096 & \color{blue}0.9776 \\
		43 & \underline{0.9999} & \underline{0.9999} & 0.9863 & 0.9986 & 0.9943 & 0.9766 & 0.9585 & \color{blue}0.9998 & \color{red}0.9997 & \color{blue}0.9998 & 0.9665 & \color{red}0.9823 & 0.9657 & 0.8976 & 0.9613 & \cellcolor{gray!40}0.7415 & 0.9544 & \underline{0.9998} & 0.9718 & \color{blue}0.9947 \\
		44 & 0.9075 & \cellcolor{gray!40}0.8984 & \cellcolor{gray!40}0.7801 & 0.9167 & \color{red}0.9240 & \cellcolor{gray!40}0.6265 & \cellcolor{gray!40}0.8837 & \underline{0.9889} & \color{blue}0.9811 & 0.9179 & 0.8666 & 0.8924 & \cellcolor{gray!40}0.7566 & 0.8523 & 0.8965 & \cellcolor{gray!40}0.2275 & 0.8785 & \underline{0.9889} & \color{blue}0.9459 & \color{red}0.9126 \\
		45 & 0.9513 & 0.9789 & 0.9249 & 0.9249 & \underline{0.9988} & \cellcolor{gray!40}0.8978 & 0.9901 & \color{blue}0.9970 & \color{red}0.9934 & 0.9877 & 0.9184 & 0.9748 & 0.9082 & 0.8599 & \color{blue}0.9933 & \cellcolor{gray!40}0.5728 & \color{red}0.9874 & \underline{0.9970} & 0.9714 & 0.9832 \\
		46 & \color{red}0.9993 & \color{blue}0.9996 & 0.9281 & 0.9885 & \cellcolor{gray!40}0.6931 & 0.9040 & 0.9949 & 0.9630 & 0.9976 & \underline{0.9997} & 0.9750 & \color{red}0.9841 & 0.9135 & 0.8543 & \cellcolor{gray!40}0.6571 & 0.8163 & \color{blue}0.9941 & 0.9630 & 0.9800 & \underline{0.9962} \\
		47 & \cellcolor{gray!40}0.8965 & \cellcolor{gray!40}0.8957 & \cellcolor{gray!40}0.8788 & \cellcolor{gray!40}0.7927 & \color{red}0.9455 & 0.9412 & \underline{0.9804} & \color{blue}0.9781 & 0.9099 & 0.9088 & \cellcolor{gray!40}0.7143 & 0.8142 & 0.8305 & \cellcolor{gray!40}0.6407 & \color{red}0.9103 & \cellcolor{gray!40}0.3525 & \color{blue}0.9682 & \underline{0.9781} & \cellcolor{gray!40}0.7584 & 0.8859 \\
		48 & 0.9712 & 0.9287 & 0.9584 & 0.9686 & 0.9092 & 0.9545 & 0.9247 & \color{blue}0.9908 & \color{red}0.9878 & \underline{0.9969} & 0.9319 & 0.8842 & 0.9342 & 0.8930 & 0.8487 & \cellcolor{gray!40}0.6863 & 0.9209 & \color{blue}0.9902 & \color{red}0.9424 & \underline{0.9915} \\
		49 & \color{blue}0.9993 & 0.9933 & 0.9583 & 0.9565 & 0.9825 & \color{red}0.9973 & \cellcolor{gray!40}0.8186 & 0.9962 & 0.9962 & \underline{0.9995} & 0.9464 & 0.9485 & 0.9434 & 0.8999 & 0.9556 & 0.8870 & 0.8166 & \underline{0.9961} & \color{red}0.9612 & \color{blue}0.9904 \\
		50 & \color{red}0.9787 & \cellcolor{gray!40}0.8547 & 0.9278 & 0.9496 & 0.9738 & \color{blue}0.9876 & \cellcolor{gray!40}0.8656 & \cellcolor{gray!40}0.8755 & 0.9724 & \underline{0.9961} & 0.8537 & \cellcolor{gray!40}0.6030 & 0.8348 & 0.8100 & \color{blue}0.8756 & \cellcolor{gray!40}0.6482 & 0.8585 & \color{red}0.8746 & 
		0.8581 & \underline{0.9819} \\
		51 & \color{red}0.9968 & 0.9966 & 0.9074 & \cellcolor{gray!40}0.8944 & \underline{0.9986} & 0.9936 & 0.9306 & 0.9957 & 0.9934 & \color{blue}0.9973 & 0.9645 & \color{red}0.9771 & 0.8974 & 0.8280 & 0.9664 & 0.8910 & 0.9292 & \underline{0.9957} & 0.9738 & \color{blue}0.9910 \\
		52 & 0.9997 & 0.9913 & 0.9551 & \underline{1.0000} & 0.9997 & \cellcolor{gray!40}0.5125 & 0.9991 & \color{red}0.9998 & \color{blue}0.9999 & 0.9988 & 0.9698 & 0.9289 & 0.8554 & 0.9423 & 0.9608 & \cellcolor{gray!40}0.1017 & \color{red}0.9898 & \underline{0.9998} & 0.9700 & \color{blue}0.9934 \\
		53 & \color{red}0.9991 & 0.9962 & 0.9971 & \cellcolor{gray!40}0.8579 & 0.9926 & \color{blue}0.9994 & \underline{0.9996} & 0.9954 & 0.9861 & 0.9970 & 0.9588 & 0.9608 & 0.9867 & \cellcolor{gray!40}0.7227 & 0.9378 & 0.8558 & \underline{0.9948} & \color{blue}0.9943 & 0.9326 & \color{red}0.9932 \\
		54 & \color{blue}0.9991 & \color{red}0.9989 & 0.9374 & \underline{0.9992} & \cellcolor{gray!40}0.6013 & 0.9779 & 0.9877 & \cellcolor{gray!40}0.6569 & 0.9925 & 0.9986 & 0.9541 & \color{red}0.9684 & 0.9120 & 0.9496 & \cellcolor{gray!40}0.5517 & 0.8788 & \color{blue}0.9848 & \cellcolor{gray!40}0.6569 & 0.9553 & \underline{0.9903} \\
		55 & \color{red}0.9643 & 0.9596 & 0.9247 & 0.9445 & \underline{0.9978} & \cellcolor{gray!40}0.6233 & 0.9606 & 0.9341 & \color{blue}0.9748 & 0.9429 & 0.8668 & 0.9072 & 0.8560 & \cellcolor{gray!40}0.7358 & \underline{0.9734} & \cellcolor{gray!40}0.2937 & \color{blue}0.9449 & \color{red}0.9341 & \cellcolor{gray!40}0.7706 
		& 0.9173 \\
		56 & \color{blue}0.9856 & 0.9470 & \cellcolor{gray!40}0.8910 & \cellcolor{gray!40}0.7882 & \underline{0.9959} & 0.9601 & 0.9645 & \cellcolor{gray!40}0.7490 & 0.9728 & \color{red}0.9832 & 0.8629 & 0.8415 & 0.8197 & \cellcolor{gray!40}0.4888 & \underline{0.9830} & \cellcolor{gray!40}0.6510 & \color{red}0.9592 & \cellcolor{gray!40}0.7490 & 0.8809 & \color{blue}0.9626 \\
		57 & \underline{0.9960} & \color{red}0.9912 & \cellcolor{gray!40}0.8498 & \cellcolor{gray!40}0.7443 & \cellcolor{gray!40}0.7134 & 0.9898 & 0.9840 & \cellcolor{gray!40}0.8921 & 0.9826 & \color{blue}0.9951 & \color{red}0.9781 & 0.9740 & 0.8314 & \cellcolor{gray!40}0.6458 & \cellcolor{gray!40}0.6499 & 0.9409 & \color{blue}0.9831 & 0.8921 & 0.9559 & \underline{0.9922} \\
		58 & \color{blue}0.9913 & 0.9826 & \cellcolor{gray!40}0.8804 & \cellcolor{gray!40}0.8740 & \cellcolor{gray!40}0.7761 & \color{red}0.9867 & 0.9843 & \cellcolor{gray!40}0.8545 & 0.9859 & \underline{0.9941} & \color{red}0.9770 & 0.9745 & 0.8609 & \cellcolor{gray!40}0.7896 & \cellcolor{gray!40}0.7165 & 0.9313 & \color{blue}0.9831 & 0.8545 & 0.9657 & \underline{0.9917} \\
		59 & \underline{1.0000} & \color{red}0.9998 & 0.9996 & 0.9997 & \color{blue}0.9999 & 0.9083 & 0.9951 & \cellcolor{gray!40}0.7706 & 0.9997 & 0.9997 & 0.9725 & \underline{0.9987} & \color{blue}0.9961 & 0.9664 & 0.9777 & \cellcolor{gray!40}0.6804 & 0.9901 & \cellcolor{gray!40}0.7706 & 0.9732 & \color{red}0.9950 \\        
		60 & 0.9539 & 0.9284 & \cellcolor{gray!40}0.8199 & 0.9053 & 0.9615 & 0.9151 & \color{blue}0.9964 & 0.9368 & \underline{0.9986} & \color{red}0.9768 & 0.9310 & 0.9251 & \cellcolor{gray!40}0.7730 & 0.8731 & 0.9184 & \cellcolor{gray!40}0.7225 & \underline{0.9901} & 0.9368 & \color{red}0.9701 & \color{blue}0.9725 \\        
		61 & 0.9923 & 0.9412 & \cellcolor{gray!40}0.7348 & 0.9895 & \cellcolor{gray!40}0.8394 & \cellcolor{gray!40}0.3755 & \color{red}0.9932 & \cellcolor{gray!40}0.7928 & \underline{0.9956} & \color{blue}0.9933 & 0.9173 & 0.8704 & \cellcolor{gray!40}0.6098 & 0.8805 & \cellcolor{gray!40}0.7940 & \cellcolor{gray!40}-0.1223 & \underline{0.9864} & \cellcolor{gray!40}0.7928 & \color{red}0.9320 & \color{blue}0.9843 \\
		62 & \color{red}0.9637 & \underline{0.9777} & \cellcolor{gray!40}0.5204 & \color{blue}0.9680 & 0.9436 & \cellcolor{gray!40}0.8062 & 0.9217 & \cellcolor{gray!40}0.8491 & 0.9593 & 0.9619 & \color{red}0.9559 & \underline{0.9758} & \cellcolor{gray!40}0.5204 & 0.9375 & 0.9273 & \cellcolor{gray!40}0.6313 & 0.9209 & 0.8491 & 
		0.9485 & \color{blue}0.9606 \\
		63 & 0.9783 & 0.9877 & \cellcolor{gray!40}0.8166 & \color{blue}0.9994 & \color{red}0.9982 & \cellcolor{gray!40}0.7913 & 0.9955 & 0.9860 & \underline{0.9998} & 0.9768 & 0.9581 & 0.9076 & \cellcolor{gray!40}0.7887 & 0.8620 & 0.9793 & \cellcolor{gray!40}0.6219 & \underline{0.9947} & \color{blue}0.9858 & \color{red}0.9815 
		& 0.9688 \\
		64 & \underline{0.9999} & \color{blue}0.9998 & 0.9450 & 0.9565 & \cellcolor{gray!40}0.8664 & \cellcolor{gray!40}0.8794 & \cellcolor{gray!40}0.8289 & 0.9963 & \color{red}0.9977 & \color{blue}0.9998 & 0.9800 & \color{blue}0.9968 & 0.9370 & 0.8923 & 0.8383 & \cellcolor{gray!40}0.6330 & 0.8255 & \color{red}0.9962 & 0.9748 
		& \underline{0.9973} \\
		65 & \underline{1.0000} & \underline{1.0000} & 0.9304 & 0.9979 & \color{blue}0.9986 & \cellcolor{gray!40}0.8606 & 0.9076 & 0.9965 & \color{red}0.9981 & \underline{1.0000} & 0.9850 & \color{red}0.9930 & 0.9264 & 0.9087 & 0.9390 & \cellcolor{gray!40}0.6008 & 0.9042 & \color{blue}0.9964 & 0.9834 & \underline{0.9977} \\   
		66 & \color{blue}0.9996 & \underline{0.9999} & \cellcolor{gray!40}0.8373 & 0.9975 & \color{red}0.9976 & \cellcolor{gray!40}0.8053 & 0.9652 & 0.9938 & 0.9969 & \underline{0.9999} & 0.9799 & \color{blue}0.9959 & 0.8355 & 0.9522 & 0.9622 & \cellcolor{gray!40}0.5315 & 0.9624 & \color{red}0.9937 & 0.9781 & \underline{0.9975} \\
		67 & 0.9669 & \color{red}0.9956 & \cellcolor{gray!40}0.7748 & 0.9580 & 0.9808 & \cellcolor{gray!40}0.4666 & 0.9356 & 0.9646 & \color{blue}0.9960 & \underline{0.9994} & 0.9362 & \color{blue}0.9928 & \cellcolor{gray!40}0.7537 & 0.8945 & 0.9654 & \cellcolor{gray!40}0.1731 & 0.9339 & 0.9642 & \color{red}0.9701 & \underline{0.9962} \\
		68 & 0.9263 & 0.9371 & \cellcolor{gray!40}0.8113 & 0.9535 & \cellcolor{gray!40}0.8776 & \cellcolor{gray!40}0.7222 & \cellcolor{gray!40}0.8291 & \color{blue}0.9932 & \color{red}0.9922 & \underline{0.9986} & 0.8948 & 0.9341 & 0.8095 & 0.9118 & 0.8658 & \cellcolor{gray!40}0.4748 & 0.8274 & \underline{0.9932} & \color{red}0.9766 & \color{blue}0.9928 \\
		69 & 0.9417 & 0.9613 & \cellcolor{gray!40}0.7419 & \color{red}0.9902 & 0.9677 & \cellcolor{gray!40}0.4714 & \cellcolor{gray!40}0.6302 & 0.9671 & \color{blue}0.9922 & \underline{0.9952} & 0.9268 & 0.9573 & \cellcolor{gray!40}0.7323 & 0.9253 & 0.9087 & \cellcolor{gray!40}0.0968 & \cellcolor{gray!40}0.6273 & \color{red}0.9656 & \color{blue}0.9772 & \underline{0.9940} \\
		70 & 0.9168 & 0.9212 & \cellcolor{gray!40}0.8312 & \color{red}0.9905 & \cellcolor{gray!40}0.8746 & \cellcolor{gray!40}0.5007 & \cellcolor{gray!40}0.7681 & 0.9555 & \color{blue}0.9952 & \underline{0.9976} & 0.8535 & 0.8634 & \cellcolor{gray!40}0.7220 & 0.8654 & 0.8231 & \cellcolor{gray!40}-0.0992 & \cellcolor{gray!40}0.7650 & \color{blue}0.9554 & \color{red}0.9264 & \underline{0.9823} \\
		71 & 0.9749 & 0.9706 & \cellcolor{gray!40}0.5180 & \color{red}0.9920 & 0.9654 & \cellcolor{gray!40}0.6223 & \cellcolor{gray!40}0.8462 & 0.9863 & \color{blue}0.9936 & \underline{0.9981} & 0.9347 & 0.9693 & \cellcolor{gray!40}0.5180 & 0.9461 & 0.9485 & \cellcolor{gray!40}0.3861 & 0.8451 & \color{blue}0.9863 & \color{red}0.9793 & \underline{0.9940} \\
		72 & 0.9265 & 0.9376 & \cellcolor{gray!40}0.8169 & 0.9293 & \cellcolor{gray!40}0.8777 & \cellcolor{gray!40}0.7459 & \cellcolor{gray!40}0.7596 & \color{blue}0.9913 & \color{red}0.9899 & \underline{0.9987} & 0.9087 & 0.9366 & 0.8159 & 0.9116 & 0.8507 & \cellcolor{gray!40}0.4949 & \cellcolor{gray!40}0.7551 & \color{blue}0.9913 & \color{red}0.9713 & \underline{0.9959} \\
		73 & 0.9666 & \cellcolor{gray!40}0.8508 & \cellcolor{gray!40}0.7562 & \color{red}0.9695 & \color{blue}0.9702 & \cellcolor{gray!40}0.5075 & \cellcolor{gray!40}0.8276 & \cellcolor{gray!40}0.7123 & 0.9394 & \underline{0.9985} & \color{red}0.9306 & 0.8395 & \cellcolor{gray!40}0.7265 & 0.8728 & \color{blue}0.9446 & \cellcolor{gray!40}0.1567 & 0.8252 & \cellcolor{gray!40}0.7121 & 0.9102 & \underline{0.9944} \\
		74 & 0.9990 & 0.9957 & 0.9834 & \underline{1.0000} & 0.9870 & \color{red}0.9991 & \underline{1.0000} & \cellcolor{gray!40}0.6347 & 0.9987 & \color{blue}0.9996 & 0.9676 & \color{red}0.9928 & 0.9581 & 0.9549 & 0.9399 & 0.8752 & \underline{0.9987} & \cellcolor{gray!40}0.6346 & 0.9532 & \color{blue}0.9962 \\
		75 & 0.9957 & \underline{0.9996} & \color{blue}0.9991 & 0.9425 & \cellcolor{gray!40}0.8992 & 0.9146 & 0.9882 & 0.9889 & \color{red}0.9991 & 0.9966 & 0.9684 & \color{blue}0.9902 & 0.9762 & 0.8132 & 0.8676 & \cellcolor{gray!40}0.7327 & 0.9771 & \color{red}0.9887 & 0.9733 & \underline{0.9929} \\
		76 & 0.9990 & \color{red}0.9998 & 0.9965 & 0.9287 & 0.9469 & 0.9935 & 0.9939 & 0.9947 & \underline{1.0000} & \color{blue}0.9999 & 0.9836 & \color{blue}0.9959 & 
		0.9808 & 0.8360 & 0.8888 & 0.8499 & 0.9892 & \color{red}0.9944 & 0.9737 & \underline{0.9983} \\
		77 & 0.9763 & 0.9373 & 0.9585 & 0.9248 & 0.9854 & \color{blue}0.9987 & \color{red}0.9971 & 0.9874 & 0.9970 & \underline{0.9999} & 0.9390 & 0.9038 & 0.8877 & \cellcolor{gray!40}0.7787 & 0.9085 & \cellcolor{gray!40}0.7952 & \color{red}0.9819 & \color{blue}0.9870 & 0.9597 & \underline{0.9965} \\
		78 & \cellcolor{gray!40}0.8551 & \cellcolor{gray!40}0.8472 & \cellcolor{gray!40}0.8340 & \underline{0.9674} & \color{blue}0.9537 & \cellcolor{gray!40}0.6876 & \cellcolor{gray!40}0.8571 & \color{red}0.9155 & 0.9052 & \cellcolor{gray!40}0.8807 & \cellcolor{gray!40}0.7686 & 0.8155 & 0.8009 & \cellcolor{gray!40}0.7659 & \underline{0.9356} & \cellcolor{gray!40}0.4612 & \color{red}0.8548 & \color{blue}0.9151 & 0.8317 & 0.8478 \\
		79 & 0.9944 & 0.9791 & 0.9410 & 0.9411 & \cellcolor{gray!40}0.7746 & \underline{0.9960} & \color{red}0.9953 & 0.9250 & \color{blue}0.9954 & 0.9948 & 0.9075 & 0.8899 & 0.8884 & 0.8247 & \cellcolor{gray!40}0.7460 & \cellcolor{gray!40}0.7335 & \underline{0.9913} & 0.9246 & \color{red}0.9319 & \color{blue}0.9811 \\        
		80 & 0.9969 & 0.9958 & \cellcolor{gray!40}0.6416 & \cellcolor{gray!40}0.8728 & 0.9942 & 0.9895 & 0.9146 & \underline{0.9994} & \color{red}0.9981 & \color{blue}0.9993 & 0.9725 & \color{red}0.9776 & \cellcolor{gray!40}0.6416 & \cellcolor{gray!40}0.7691 & 0.9699 & 0.8748 & 0.9112 & \underline{0.9993} & 0.9604 & \color{blue}0.9864 \\
		81 & 0.9929 & 0.9663 & 0.9699 & 0.9711 & 0.9942 & 0.9338 & 0.9726 & \underline{0.9968} & \color{blue}0.9965 & \color{red}0.9954 & 0.9718 & 0.9653 & 0.9671 & 0.9382 & \color{red}0.9830 & \cellcolor{gray!40}0.7193 & 0.9671 & \underline{0.9967} & 0.9742 & \color{blue}0.9876 \\
		82 & \cellcolor{gray!40}0.8677 & \underline{0.9855} & \cellcolor{gray!40}0.4996 & \cellcolor{gray!40}0.6092 & \color{blue}0.9085 & \cellcolor{gray!40}0.5950 & \cellcolor{gray!40}0.8620 & \cellcolor{gray!40}0.7250 & \cellcolor{gray!40}0.8736 & \cellcolor{gray!40}0.8501 & 0.8367 & \underline{0.9779} & \cellcolor{gray!40}0.4752 & \cellcolor{gray!40}0.3304 & \color{blue}0.8942 & \cellcolor{gray!40}0.4190 & \color{red}0.8582 & \cellcolor{gray!40}0.7250 & 0.8362 & 0.8195 \\        
		83 & \color{blue}0.9920 & 0.9912 & \cellcolor{gray!40}0.6805 & \cellcolor{gray!40}0.5545 & \cellcolor{gray!40}0.8842 & \underline{0.9939} & 0.9215 & \cellcolor{gray!40}0.8992 & 0.9803 & \color{red}0.9919 & \color{red}0.9860 & \underline{0.9909} & \cellcolor{gray!40}0.6802 & \cellcolor{gray!40}0.4957 & 0.8724 & 0.9606 & 0.9205 & 0.8992 & 0.9677 & \color{blue}0.9886 \\
		84 & \underline{0.9994} & \underline{0.9994} & 0.9703 & 0.9905 & \color{blue}0.9993 & 0.9934 & 0.9595 & 0.9966 & 0.9969 & \color{red}0.9981 & 0.9780 & \underline{0.9975} & 0.9670 & 0.9686 & 0.9790 & 0.8862 & 0.9564 & \color{blue}0.9963 & 0.9672 & \color{red}0.9907 \\
		85 & \cellcolor{gray!40}0.8242 & \cellcolor{gray!40}0.6585 & \cellcolor{gray!40}0.7006 & \cellcolor{gray!40}0.8319 & \color{red}0.9488 & \cellcolor{gray!40}0.8589 & \cellcolor{gray!40}0.8269 & \underline{0.9871} & \color{blue}0.9604 & 0.9294 & \cellcolor{gray!40}0.7578 & \cellcolor{gray!40}0.5742 & \cellcolor{gray!40}0.6395 & \cellcolor{gray!40}0.6078 & \color{blue}0.9368 & \cellcolor{gray!40}0.6993 & 0.8205 & \underline{0.9869} & 0.8752 & \color{red}0.9091 \\
		86 & \color{red}0.9936 & 0.9911 & 0.9445 & 0.9927 & 0.9533 & 0.9888 & \cellcolor{gray!40}0.8959 & \cellcolor{gray!40}0.4797 & \color{blue}0.9955 & \underline{0.9966} & 0.9511 & 0.9591 & 0.9245 & \color{blue}0.9732 & 0.9041 & 0.8881 & 0.8802 & \cellcolor{gray!40}0.4797 & \color{red}0.9615 & \underline{0.9777} \\        
		87 & \color{blue}0.9998 & \color{blue}0.9998 & \cellcolor{gray!40}0.8285 & \color{red}0.9997 & \cellcolor{gray!40}0.6590 & \underline{0.9999} & 0.9962 & 0.9988 
		& 0.9958 & \color{red}0.9997 & 0.9893 & \color{red}0.9980 & 0.8064 & 0.9908 & \cellcolor{gray!40}0.5932 & 0.9745 & 0.9960 & \underline{0.9988} & 0.9864 & \color{blue}0.9986 \\
		88 & 0.9863 & 0.9707 & 0.9223 & \cellcolor{gray!40}0.5598 & 0.9731 & \color{red}0.9954 & 0.9801 & \underline{0.9966} & 0.9892 & \color{blue}0.9960 & 0.9632 & 0.9624 & 0.9155 & \cellcolor{gray!40}0.4933 & 0.9531 & \cellcolor{gray!40}0.7876 & \color{red}0.9795 & \underline{0.9965} & 0.9692 & \color{blue}0.9940 \\        
		89 & \color{blue}0.9999 & 0.9990 & 0.9921 & \underline{1.0000} & \color{red}0.9997 & \underline{1.0000} & 0.9847 & \color{blue}0.9999 & 0.9975 & \color{blue}0.9999 & 0.9858 & \color{red}0.9921 & 0.9872 & 0.9858 & 0.9753 & 0.8610 & 0.9834 & \underline{0.9999} & 0.9830 & \color{blue}0.9971 \\
		90 & \underline{1.0000} & \underline{1.0000} & 0.9918 & \underline{1.0000} & \color{red}0.9998 & \color{blue}0.9999 & \color{red}0.9998 & 0.9627 & 0.9961 & \underline{1.0000} & 0.9827 & \color{red}0.9872 & 0.9840 & 0.9620 & 0.9788 & 0.8823 & \underline{0.9991} & 0.9624 & 0.9742 & \color{blue}0.9976 \\
		91 & \color{red}0.9954 & 0.9932 & 0.9687 & \cellcolor{gray!40}0.6192 & \color{blue}0.9963 & 0.9839 & 0.9605 & \underline{0.9993} & 0.9837 & 0.9912 & 0.9709 & 0.9708 & 0.9577 & \cellcolor{gray!40}0.4992 & \color{red}0.9839 & \cellcolor{gray!40}0.7634 & 0.9594 & \underline{0.9993} & 0.9559 & \color{blue}0.9856 \\        
		92 & \color{blue}0.9947 & 0.9873 & \cellcolor{gray!40}0.8795 & \cellcolor{gray!40}0.8958 & 0.9880 & 0.9843 & 0.9509 & 0.9637 & \color{red}0.9920 & \underline{0.9952} & \color{blue}0.9763 & \color{red}0.9751 & 0.8709 & 0.8557 & 0.9612 & 0.8367 & 0.9492 & 0.9635 & 0.9703 & \underline{0.9905} \\
		93 & \color{blue}0.9998 & \underline{0.9999} & 0.9926 & 0.9278 & 0.9671 & 0.9826 & 0.9995 & 0.9993 & 0.9995 & \color{red}0.9996 & 0.9796 & 0.9939 & 0.9822 & \cellcolor{gray!40}0.7718 & 0.9628 & \cellcolor{gray!40}0.7022 & \color{blue}0.9982 & \underline{0.9993} & 0.9651 & \color{red}0.9970 \\
		94 & \color{red}0.9918 & \underline{0.9994} & \cellcolor{gray!40}0.8670 & \cellcolor{gray!40}0.7902 & 0.9708 & \cellcolor{gray!40}0.3763 & \cellcolor{gray!40}0.8937 & 0.9814 & \cellcolor{gray!40}0.8543 & \color{blue}0.9956 & 0.9806 & \underline{0.9944} & 0.8370 & \cellcolor{gray!40}0.6413 & 0.9298 & \cellcolor{gray!40}0.0029 & 0.8892 & \color{red}0.9813 & \cellcolor{gray!40}0.7977 & \color{blue}0.9937 \\ \midrule
		Avg. & \color{red}0.9800  & 0.9730  & 0.8672  & 0.9284  & 0.9418  & 0.8715  & 0.9426  & 0.9409  & \color{blue}0.9877  & \underline{0.9899}  & 0.9416  & \color{red}0.9436  & 0.8453  & 0.8458  & 0.9076  & 0.6587  & 0.9391  & 0.9408  & \color{blue}0.9538  & \underline{0.9836} \\ \bottomrule[0.5mm]
	\end{tabular}
\label{t:1}
\end{table*}

\begin{table}
	\centering
	\caption{Performance of Various Methods on 94 HSIs from the HAD100 Dataset is Evaluated Based on the Number of Best, Top two, Top three, and Failure Results Achieved. The Best Result is Highlighted.}
	\tiny
	\tabcolsep=0.15cm
	\begin{tabular}{c|cccccccccc}
		\toprule[0.5mm]
		\multicolumn{11}{c}{$AUC_{(D,F)}$} \\ \midrule
		Number of & RX & LRX & CRD & LRASR & AED & PTA & Auto-AD & DFAE & AETNet & \textbf{STAD} \\ \midrule
		Top-1 & 15 & 17 & 0 &  8 &  8 &  6 & 4 & 15 &  5 & \textbf{37}  \\
		Top-2 & 34 & 24 & 3 & 14 & 16 & 11 & 5 & 26 & 25 & \textbf{59}  \\
		Top-3 & 48 & 32 & 3 & 21 & 27 & 20 & 13 & 35 & 43 & \textbf{72}  \\ 
		Failure & 5 & 6 & 42 & 20 & 18 & 33 & 19 & 18 & \textbf{2} & \textbf{2}  \\ 
		\midrule
		\multicolumn{11}{c}{$AUC_{BS}$} \\ \midrule
		Number of & RX & LRX & CRD & LRASR & AED & PTA & Auto-AD & DFAE & AETNet & \textbf{STAD} \\ \midrule
		Top-1 & 0 & 9 & 0 & 0 &  3 & 0 & 9 & \textbf{42} & 0 & 32 \\
		Top-2 & 2 & 23 & 3 & 1 &  8 & 0 & 19 & 54 & 5 & \textbf{74} \\
		Top-3 & 9 & 41 & 6 & 1 & 14 & 0 & 35 & 66 & 26 & \textbf{85} \\
		Failure & 4 & 2 & 25 & 20 & 10 & 60 & 4 & 11 & 3 & \textbf{0} \\ 
		\bottomrule[0.5mm]
	\end{tabular}
\label{t:2}
\end{table}

\subsection{Ablation Study}
In this section, we conduct two sets of experiments to validate the effectiveness of different components of the proposed method on HAD. The first set of experiments aimed to demonstrate the feasibility, validity, and robustness of solving the HAD problem in feature space. In detail, the following five scenarios are designed: (A) We directly use the reconstruction error as the detection result. (B) We directly use the saliency map as the detection result. (C) We directly use the small target mask matrix output from STF as the detection result. (D) We multiply the reconstruction error and the small target mask matrix as the detection result. (E) We consider the complete proposed method STAD. These scenarios are designed to explore and emphasize different aspects of the proposed method, demonstrating its effectiveness and generality in solving HAD problems. 

As shown in Table \ref{t:3}, when solely relying on reconstruction errors, both average detection accuracy and background suppression perform poorly. This aligns with our earlier assertion that anomalies, which are easily distinguishable to the human eye, may assume a subtle representation when observed within the reconstructed error space. The saliency map is a more accurate detection result than the reconstruction error. This suggests that the feature space retains more anomalous representations but lacks more specific task guidance. STF provides guidance for solving problems in the feature space. Although it yields high detection metrics when used directly as a detection result, its visualization results reveal more background pixels. Therefore, STF is only a suitable guide for deep learning methods. When combined with the reconstruction error, it produces better background suppression results. However, the feature space appears to be more robust than the reconstructed error space. STAD utilizes the STF-generated mask matrix to guide the solution in the feature space by masking the activation of low-confidence neurons. As a result, STAD achieves significantly improved results compared to other scenarios.

Another set of experiments is designed to validate the feasibility and effectiveness of our proposed teacher and student networks and distillation strategy. In detail, the following four scenarios are designed: (A) We use the teacher network directly for anomaly detection. (B) We use the student network directly for anomaly detection. (C) We distill the knowledge learned from the teacher network into a simple fully connected layer network. (D) We consider the complete proposed method STAD, which distill the knowledge learned by the teacher network into the proposed student network.

As shown in Table \ref{t:4}, applying only teacher network inference consumes a lot of computational and memory resources. Additionally, the teacher network is overly complex and contains redundant parameters, which can generate noise during backpropagation and thus affect the detection performance. However, solely training the student network results in a failure to acquire additional HSI features, even though its throughput and memory usage are better than the teacher network. These two scenarios demonstrate the necessity and effectiveness of using knowledge distillation in HAD tasks. Additionally, we designed a fully connected layer network with the same number of layers as our proposed student network. The experimental results demonstrate that using only a fully connected layer network results in a loss of spatial information, which leads to a degradation of the detection performance. 

Overall, each component of the STAD model is designed with the HAD task as a guide and provides successful innovation in the solution space of the problem. This contributes significantly to the development of HAD technologies.

\begin{table}[t]
	\centering
	\tabcolsep=0.5cm
	\caption{Average Results of Ablation Study for STAD Performed on HAD100 Dataset. The Best Result is Highlighted.}
	\begin{tabular}{c|cc}
		\toprule[0.5mm]
		Description & $AUC_{(D,F)}$ & $AUC_{BS}$ \\ \midrule
		(A) & 0.7827 & 0.5885 \\
		(B) & 0.8774 & 0.6906 \\
		(C) & 0.9891 & 0.9588 \\
		(D) & 0.9894 & 0.9646 \\
		(E) & \textbf{0.9899} & \textbf{0.9836} \\
		\bottomrule[0.5mm]
	\end{tabular}
	\label{t:3}
\end{table}

\begin{table}[ht]
	\centering
	\caption{Average Results of Ablation Study for STAD Performed on HAD100 Dataset. Throughput and Memory Usage are Calculated Only for the Inference Process. Due to the Idea of Pixel-level Detection in HAD, Throughput Refers to the Number of Pixels Processed Per Second. The Best Result is Highlighted.}
	\begin{tabular}{c|cccc}
		\toprule[0.5mm]
		Description & $AUC_{(D,F)}$ & $AUC_{BS}$ & Throughput & Memory usage \\ \midrule
		(A) & 0.9896 & 0.9830 & 0.28 MPS & 7445 MB \\
		(B) & 0.9874 & 0.9824 & 1.09 MPS & 2157 MB \\
		(C) & 0.9895 & 0.9816 & \textbf{1.29} MPS & \textbf{2091} MB \\
		(D) & \textbf{0.9899} & \textbf{0.9836} & 1.09 MPS & 2157 MB \\
		\bottomrule[0.5mm]
	\end{tabular}
	\label{t:4}
\end{table}

\subsection{Dependency Study}
To enhance detection accuracy, many deep learning-based methods for HAD combine traditional anomaly detectors, such as WeaklyAD preprocessing the image by coarse inspection and DFAE using AED for post-processing. This combination of data-driven and model-driven technique can leverage artificial prior knowledge to solve some problems that are difficult for deep learning. Therefore, in STAD, the global Mahalanobis distance is utilized to flag small targets, in order to avoid the neural network's attention being drawn towards non-anomalous large targets. However, the use of traditional detectors presents dependency issues. Typically, these detectors require specific hyperparameters to be set for particular scenes. And if deep learning-based approaches become overly dependent on the detectors utilized, they may hinder the generalization of HAD methods and fail to operate in some HSIs.

\begin{figure}[!]
	\centering
	\includegraphics[width=0.5\textwidth]{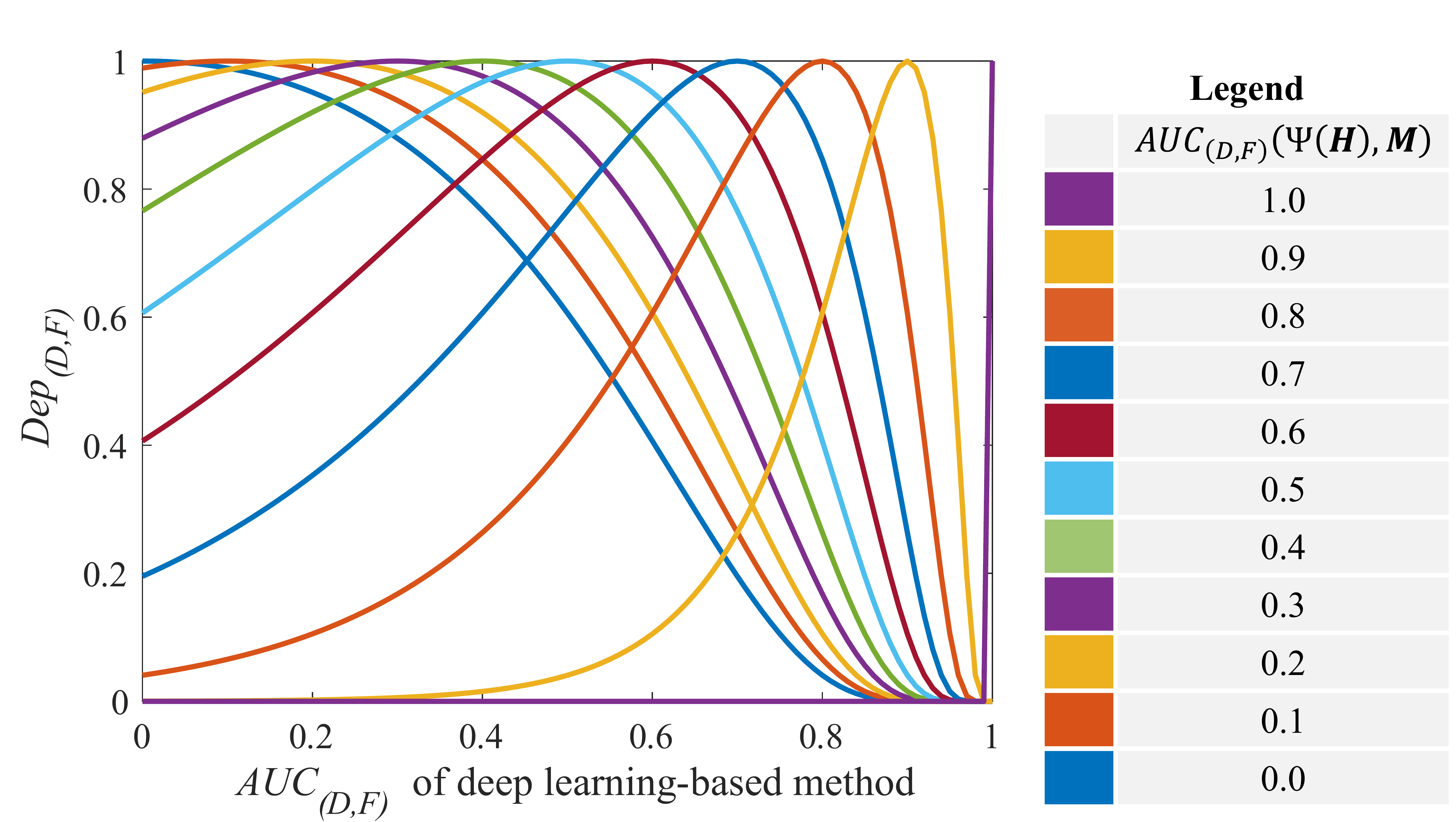}
	\caption{Dependency curves in different situations.}
	\label{fig:7}
\end{figure}

Therefore, in this section, we study the dependence of STAD on traditional methods. This is also a tolerance analysis of STAD against traditional methods, showcasing the robustness inherent in the STAD approach. To quantify dependence, we formulated an indicator called $Dep_{(\cdot)}(\cdot)$ based on the tolerance of multiple regression analysis in statistics. Given the detector $\Phi(\cdot)$, and the traditional method $\Psi(\cdot)$ used by it, $Dep_{(\cdot)}(\cdot)$ is calculated as

\begin{multline}
	\label{dep}
	Dep_{(D,F)}(\Phi, \Psi, \boldsymbol{H}, \boldsymbol{M}) = \\
	\frac{1}{e^{\frac{(AUC_{(D,F)}(\Phi(\boldsymbol{H}),\boldsymbol{M}) - AUC_{(D,F)}(\Psi(\boldsymbol{H}),\boldsymbol{M}))^2}{(1 - AUC_{(D,F)}(\Phi(\boldsymbol{H}),\boldsymbol{M}) + \beta)(1 - AUC_{(D,F)}(\Psi(\boldsymbol{H}),\boldsymbol{M}) + \beta)}}},
\end{multline}
where $\boldsymbol{H}$ denotes the test HSI, $\boldsymbol{M}$ denotes the anomaly map corresponding to the test HSI, and $\beta$ denotes the bias. The value of $\beta$ is set to $10^{-12}$. This metric is designed with various considerations in mind. Here, $(AUC_{(D,F)}(\Phi(\boldsymbol{H}),\boldsymbol{M}) - AUC_{(D,F)}(\Psi(\boldsymbol{H}),\boldsymbol{M}))^2$ represents the difference in performance between deep learning-based method and traditional method, and a higher value for this parameter indicates a lesser dependency. $(1 - AUC_{(D,F)}(\Phi(\boldsymbol{H}),\boldsymbol{M}) + \beta)(1 - AUC_{(D,F)}(\Psi(\boldsymbol{H}),\boldsymbol{M}) + \beta)$ takes into account the performance of both methods. In situations where the detection difficulty of HSI is low, and the $AUC_{(D,F)}$ of both methods is close to 1, the performance difference becomes challenging to reflect the dependency. Therefore, this item serves as a scaling factor to amplify the difference in such scenarios. The role of parameter $\beta$ is to prevent division by zero, ensuring the validity of the operation. The exponent serves the purpose of adjusting the domain of values, ensuring that the dependence metric falls within the range (0,1]. For a more intuitive understanding of $Dep_{(D,F)}(\cdot)$, we present dependency curves for different cases, as depicted in Figure. \ref{fig:7}. It is evident that the dependency is highest when the performance of both methods is equal. In a similar fashion, the dependence on background suppression ability can be mathematically expressed as:

\begin{multline}
	\label{dep}
	Dep_{BS}(\Phi, \Psi, \boldsymbol{H}, \boldsymbol{M}) = \\
	\frac{1}{e^{\frac{(AUC_{BS}(\Phi(\boldsymbol{H}),\boldsymbol{M}) - AUC_{BS}(\Psi(\boldsymbol{H}),\boldsymbol{M}))^2}{(1 - AUC_{BS}(\Phi(\boldsymbol{H}),\boldsymbol{M}) + \beta)(1 - AUC_{BS}(\Psi(\boldsymbol{H}),\boldsymbol{M}) + \beta)}}},
\end{multline}

A higher dependency suggests that the respective deep learning-based method is more influenced by traditional methods and exhibits poorer generalization. Conversely, lower dependency indicates higher fault tolerance and robustness. Building upon this, we conducted an analysis of the dependence of the proposed STAD on traditional methods.

\begin{figure}[!]
	\centering
	\includegraphics[width=0.5\textwidth]{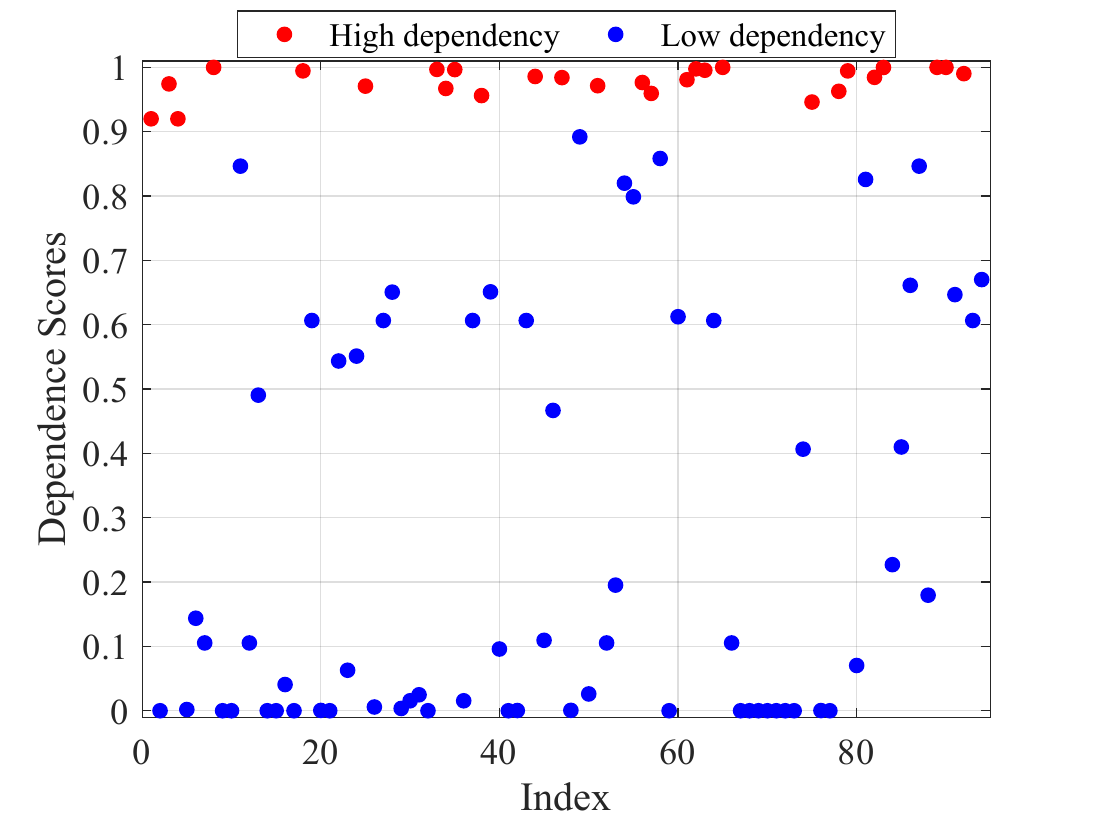}
	\caption{Scatterplot of STAD dependence scores on traditional methods on the HAD100 dataset.}
	\label{fig:8}
\end{figure}

We present the dependence scores on each test HSI concerning the global Mahalanobis distance and STAD, as illustrated in Figure \ref{fig:8}. We define dependency scores greater than 0.9 as indicating strong dependencies. On the HAD100 dataset, the percentage of STAD that exhibits strong dependence on traditional methods is 28.72\%, with an average dependency score $mDep_{(D,F)}$ of 0.4698, where $mDep_{(D,F)}$ is calculated as

\begin{equation}
	mDep_{(D,F)} = \frac{\sum_{i}^{I} R^i Dep_{(D,F)}(\Phi, \Psi, \boldsymbol{H}^i, \boldsymbol{M}^i)}{\sum_{i}^{I} R^i},
\end{equation}
where $I$ represents the total number of test HSIs, and $R^i$ denotes the covariance of the performance sequences of the two methods after the removal of the $i$-th test HSI. In summary, STAD exhibits lower dependency on traditional methods, showcasing high fault tolerance and robustness.

\section{Conclusion}
The majority of existing HAD methods rely on reconstruction, statistics, and representation, lacking the capability to directly detect anomalies from a feature space with robust visual perception. In this paper, we introduce a novel small target-aware detector named STAD, which strategically transfers the solution space of HAD to the feature space, aiming to enhance both detection performance and robustness. We illustrate that the feature space provides a more intricate representation of anomalies, aligning with human visual perception. In particular, we introduce the saliency map to capture anomalous representations within the feature space. Additionally, a small-target filter is devised to selectively mask the activation of low-confidence neurons. STAD presents a novel approach and methodology for addressing the HAD problem, which is highly valuable for research purposes. We validate our method on the HAD100 dataset, and the experimental results unequivocally indicate that STAD achieves the highest levels of both detection performance and robustness.

\newpage
\bibliography{ref}

\begin{thebibliography}{10}
\providecommand{\url}[1]{#1}
\csname url@samestyle\endcsname
\providecommand{\newblock}{\relax}
\providecommand{\bibinfo}[2]{#2}
\providecommand{\BIBentrySTDinterwordspacing}{\spaceskip=0pt\relax}
\providecommand{\BIBentryALTinterwordstretchfactor}{4}
\providecommand{\BIBentryALTinterwordspacing}{\spaceskip=\fontdimen2\font plus
\BIBentryALTinterwordstretchfactor\fontdimen3\font minus
  \fontdimen4\font\relax}
\providecommand{\BIBforeignlanguage}[2]{{%
\expandafter\ifx\csname l@#1\endcsname\relax
\typeout{** WARNING: IEEEtran.bst: No hyphenation pattern has been}%
\typeout{** loaded for the language `#1'. Using the pattern for}%
\typeout{** the default language instead.}%
\else
\language=\csname l@#1\endcsname
\fi
#2}}
\providecommand{\BIBdecl}{\relax}
\BIBdecl

\bibitem{hsi}
M.~Borengasser, W.~S. Hungate, and R.~Watkins, \emph{Hyperspectral remote
  sensing: principles and applications}.\hskip 1em plus 0.5em minus 0.4em\relax
  CRC press, 2007.

\bibitem{nongye-hsi}
B.~Datt, T.~R. McVicar, T.~G. Van~Niel, D.~L. Jupp, and J.~S. Pearlman,
  ``Preprocessing eo-1 hyperion hyperspectral data to support the application
  of agricultural indexes,'' \emph{IEEE Transactions on Geoscience and Remote
  Sensing}, vol.~41, no.~6, pp. 1246--1259, 2003.

\bibitem{kuangchan-hsi}
B.~H{\"o}rig, F.~K{\"u}hn, F.~Osch{\"u}tz, and F.~Lehmann, ``Hymap
  hyperspectral remote sensing to detect hydrocarbons,'' \emph{International
  Journal of Remote Sensing}, vol.~22, no.~8, pp. 1413--1422, 2001.

\bibitem{zaihai-hsi}
N.~Zhang, G.~Yang, Y.~Pan, X.~Yang, L.~Chen, and C.~Zhao, ``A review of
  advanced technologies and development for hyperspectral-based plant disease
  detection in the past three decades,'' \emph{Remote Sensing}, vol.~12,
  no.~19, p. 3188, 2020.

\bibitem{jiuyuan-hsi}
M.~T. Eismann, A.~D. Stocker, and N.~M. Nasrabadi, ``Automated hyperspectral
  cueing for civilian search and rescue,'' \emph{Proceedings of the IEEE},
  vol.~97, no.~6, pp. 1031--1055, 2009.

\bibitem{had}
L.~Li, W.~Li, Q.~Du, and R.~Tao, ``Low-rank and sparse decomposition with
  mixture of gaussian for hyperspectral anomaly detection,'' \emph{IEEE
  Transactions on Cybernetics}, vol.~51, no.~9, pp. 4363--4372, 2020.

\bibitem{hc}
X.~Yang, W.~Cao, Y.~Lu, and Y.~Zhou, ``{QTN}: Quaternion transformer network
  for hyperspectral image classification,'' \emph{IEEE Transactions on Circuits
  and Systems for Video Technology}, vol.~33, no.~12, pp. 7370--7384, 2023.

\bibitem{tcsvt2}
M.~Li, Y.~Liu, G.~Xue, Y.~Huang, and G.~Yang, ``Exploring the relationship
  between center and neighborhoods: Central vector oriented self-similarity
  network for hyperspectral image classification,'' \emph{IEEE Transactions on
  Circuits and Systems for Video Technology}, vol.~33, no.~4, pp. 1979--1993,
  2023.

\bibitem{tcsvt3}
J.~Fan, T.~Chen, and S.~Lu, ``Superpixel guided deep-sparse-representation
  learning for hyperspectral image classification,'' \emph{IEEE Transactions on
  Circuits and Systems for Video Technology}, vol.~28, no.~11, pp. 3163--3173,
  2018.

\bibitem{tcsvt4}
J.~Xie, N.~He, L.~Fang, and P.~Ghamisi, ``Multiscale densely-connected fusion
  networks for hyperspectral images classification,'' \emph{IEEE Transactions
  on Circuits and Systems for Video Technology}, vol.~31, no.~1, pp. 246--259,
  2021.

\bibitem{ldx}
D.~Li, W.~Xie, Y.~Li, and L.~Fang, ``Fedfusion: Manifold driven federated
  learning for multi-satellite and multi-modality fusion,'' \emph{arXiv
  preprint arXiv:2311.09540}, 2023.

\bibitem{tcsvt5}
W.~Dong, T.~Yang, J.~Qu, T.~Zhang, S.~Xiao, and Y.~Li, ``Joint contextual
  representation model-informed interpretable network with dictionary aligning
  for hyperspectral and lidar classification,'' \emph{IEEE Transactions on
  Circuits and Systems for Video Technology}, vol.~33, no.~11, pp. 6804--6818,
  2023.

\bibitem{htd}
N.~M. Nasrabadi, ``Hyperspectral target detection: An overview of current and
  future challenges,'' \emph{IEEE Signal Processing Magazine}, vol.~31, no.~1,
  pp. 34--44, 2013.

\bibitem{rx}
I.~S. Reed and X.~Yu, ``Adaptive multiple-band cfar detection of an optical
  pattern with unknown spectral distribution,'' \emph{IEEE Transactions on
  Acoustics, Speech, and Signal Processing}, vol.~38, no.~10, pp. 1760--1770,
  1990.

\bibitem{lrx}
J.~M. Molero, E.~M. Garzon, I.~Garcia, and A.~Plaza, ``Analysis and
  optimizations of global and local versions of the rx algorithm for anomaly
  detection in hyperspectral data,'' \emph{IEEE Journal of Selected Topics in
  Applied Earth Observations and Remote Sensing}, vol.~6, no.~2, pp. 801--814,
  2013.

\bibitem{qlrx}
C.~E. Caefer, J.~Silverman, O.~Orthal, D.~Antonelli, Y.~Sharoni, and S.~R.
  Rotman, ``Improved covariance matrices for point target detection in
  hyperspectral data,'' \emph{Optical Engineering}, vol.~47, no.~7, pp.
  076\,402--076\,402, 2008.

\bibitem{lairx}
Y.~P. Taitano, B.~A. Geier, and K.~W. Bauer, ``A locally adaptable iterative rx
  detector,'' \emph{EURASIP Journal on Advances in Signal Processing}, vol.
  2010, pp. 1--10, 2010.

\bibitem{crd}
W.~Li and Q.~Du, ``Collaborative representation for hyperspectral anomaly
  detection,'' \emph{IEEE Transactions on Geoscience and Remote Sensing},
  vol.~53, no.~3, pp. 1463--1474, 2014.

\bibitem{swcrd}
R.~Wang, H.~Hu, F.~He, F.~Nie, S.~Cai, and Z.~Ming, ``Self-weighted
  collaborative representation for hyperspectral anomaly detection,''
  \emph{Signal Processing}, vol. 177, p. 107718, 2020.

\bibitem{srd}
J.~Li, H.~Zhang, L.~Zhang, and L.~Ma, ``Hyperspectral anomaly detection by the
  use of background joint sparse representation,'' \emph{IEEE Journal of
  Selected Topics in Applied Earth Observations and Remote Sensing}, vol.~8,
  no.~6, pp. 2523--2533, 2015.

\bibitem{lrasr}
Y.~Xu, Z.~Wu, J.~Li, A.~Plaza, and Z.~Wei, ``Anomaly detection in hyperspectral
  images based on low-rank and sparse representation,'' \emph{IEEE Transactions
  on Geoscience and Remote Sensing}, vol.~54, no.~4, pp. 1990--2000, 2015.

\bibitem{lsdm}
L.~Li, W.~Li, Q.~Du, and R.~Tao, ``Low-rank and sparse decomposition with
  mixture of gaussian for hyperspectral anomaly detection,'' \emph{IEEE
  Transactions on Cybernetics}, vol.~51, no.~9, pp. 4363--4372, 2020.

\bibitem{aed}
X.~Kang, X.~Zhang, S.~Li, K.~Li, J.~Li, and J.~A. Benediktsson, ``Hyperspectral
  anomaly detection with attribute and edge-preserving filters,'' \emph{IEEE
  Transactions on Geoscience and Remote Sensing}, vol.~55, no.~10, pp.
  5600--5611, 2017.

\bibitem{stdg}
W.~Xie, T.~Jiang, Y.~Li, X.~Jia, and J.~Lei, ``Structure tensor and guided
  filtering-based algorithm for hyperspectral anomaly detection,'' \emph{IEEE
  Transactions on Geoscience and Remote Sensing}, vol.~57, no.~7, pp.
  4218--4230, 2019.

\bibitem{tdad}
X.~Zhang, G.~Wen, and W.~Dai, ``A tensor decomposition-based anomaly detection
  algorithm for hyperspectral image,'' \emph{IEEE Transactions on Geoscience
  and Remote Sensing}, vol.~54, no.~10, pp. 5801--5820, 2016.

\bibitem{tpca}
Z.~Chen, B.~Yang, and B.~Wang, ``A preprocessing method for hyperspectral
  target detection based on tensor principal component analysis,'' \emph{Remote
  Sensing}, vol.~10, no.~7, p. 1033, 2018.

\bibitem{pta}
L.~Li, W.~Li, Y.~Qu, C.~Zhao, R.~Tao, and Q.~Du, ``Prior-based tensor
  approximation for anomaly detection in hyperspectral imagery,'' \emph{IEEE
  Transactions on Neural Networks and Learning Systems}, vol.~33, no.~3, pp.
  1037--1050, 2020.

\bibitem{frfe}
R.~Tao, X.~Zhao, W.~Li, H.-C. Li, and Q.~Du, ``Hyperspectral anomaly detection
  by fractional fourier entropy,'' \emph{IEEE Journal of Selected Topics in
  Applied Earth Observations and Remote Sensing}, vol.~12, no.~12, pp.
  4920--4929, 2019.

\bibitem{ssdf}
S.~Chang, B.~Du, and L.~Zhang, ``A subspace selection-based discriminative
  forest method for hyperspectral anomaly detection,'' \emph{IEEE Transactions
  on Geoscience and Remote Sensing}, vol.~58, no.~6, pp. 4033--4046, 2020.

\bibitem{ae}
G.~E. Hinton and R.~Zemel, ``Autoencoders, minimum description length and
  helmholtz free energy,'' \emph{Advances in Neural Information Processing
  Systems}, vol.~6, 1993.

\bibitem{aae}
A.~Makhzani, J.~Shlens, N.~Jaitly, I.~Goodfellow, and B.~Frey, ``Adversarial
  autoencoders,'' \emph{arXiv preprint arXiv:1511.05644}, 2015.

\bibitem{gan}
Y.~Hong, U.~Hwang, J.~Yoo, and S.~Yoon, ``How generative adversarial networks
  and their variants work: An overview,'' \emph{ACM Computing Surveys (CSUR)},
  vol.~52, no.~1, pp. 1--43, 2019.

\bibitem{safl}
W.~Xie, B.~Liu, Y.~Li, J.~Lei, C.-I. Chang, and G.~He, ``Spectral adversarial
  feature learning for anomaly detection in hyperspectral imagery,'' \emph{IEEE
  Transactions on Geoscience and Remote Sensing}, vol.~58, no.~4, pp.
  2352--2365, 2019.

\bibitem{sdlr}
J.~Lei, S.~Fang, W.~Xie, Y.~Li, and C.-I. Chang, ``Discriminative
  reconstruction for hyperspectral anomaly detection with spectral learning,''
  \emph{IEEE Transactions on Geoscience and Remote Sensing}, vol.~58, no.~10,
  pp. 7406--7417, 2020.

\bibitem{hadgan}
T.~Jiang, Y.~Li, W.~Xie, and Q.~Du, ``Discriminative reconstruction constrained
  generative adversarial network for hyperspectral anomaly detection,''
  \emph{IEEE Transactions on Geoscience and Remote Sensing}, vol.~58, no.~7,
  pp. 4666--4679, 2020.

\bibitem{tdcnn}
W.~Li, G.~Wu, and Q.~Du, ``Transferred deep learning for anomaly detection in
  hyperspectral imagery,'' \emph{IEEE Geoscience and Remote Sensing Letters},
  vol.~14, no.~5, pp. 597--601, 2017.

\bibitem{autoad}
S.~Wang, X.~Wang, L.~Zhang, and Y.~Zhong, ``Auto-ad: Autonomous hyperspectral
  anomaly detection network based on fully convolutional autoencoder,''
  \emph{IEEE Transactions on Geoscience and Remote Sensing}, vol.~60, pp.
  1--14, 2021.

\bibitem{rgae}
G.~Fan, Y.~Ma, X.~Mei, F.~Fan, J.~Huang, and J.~Ma, ``Hyperspectral anomaly
  detection with robust graph autoencoders,'' \emph{IEEE Transactions on
  Geoscience and Remote Sensing}, vol.~60, pp. 1--14, 2021.

\bibitem{e2e}
K.~Jiang, W.~Xie, J.~Lei, Z.~Li, Y.~Li, T.~Jiang, and Q.~Du, ``E2e-liade:
  End-to-end local invariant autoencoding density estimation model for anomaly
  target detection in hyperspectral image,'' \emph{IEEE Transactions on
  Cybernetics}, vol.~52, no.~11, pp. 11\,385--11\,396, 2021.

\bibitem{lren}
K.~Jiang, W.~Xie, J.~Lei, T.~Jiang, and Y.~Li, ``Lren: Low-rank embedded
  network for sample-free hyperspectral anomaly detection,'' in
  \emph{Proceedings of the AAAI Conference on Artificial Intelligence},
  vol.~35, no.~5, 2021, pp. 4139--4146.

\bibitem{wad}
T.~Jiang, W.~Xie, Y.~Li, J.~Lei, and Q.~Du, ``Weakly supervised discriminative
  learning with spectral constrained generative adversarial network for
  hyperspectral anomaly detection,'' \emph{IEEE Transactions on Neural Networks
  and Learning Systems}, vol.~33, no.~11, pp. 6504--6517, 2021.

\bibitem{dfae}
Y.~Liu, W.~Xie, Y.~Li, Z.~Li, and Q.~Du, ``Dual-frequency autoencoder for
  anomaly detection in transformed hyperspectral imagery,'' \emph{IEEE
  Transactions on Geoscience and Remote Sensing}, vol.~60, pp. 1--13, 2022.

\bibitem{aetnet}
Z.~Li, Y.~Wang, C.~Xiao, Q.~Ling, Z.~Lin, and W.~An, ``You only train once:
  Learning a general anomaly enhancement network with random masks for
  hyperspectral anomaly detection,'' \emph{IEEE Transactions on Geoscience and
  Remote Sensing}, vol.~61, pp. 1--18, 2023.

\bibitem{feature}
R.~Zhang, P.~Isola, A.~A. Efros, E.~Shechtman, and O.~Wang, ``The unreasonable
  effectiveness of deep features as a perceptual metric,'' in \emph{Proceedings
  of the IEEE Conference on Computer Vision and Pattern Recognition}, 2018, pp.
  586--595.

\bibitem{saliency}
K.~Simonyan, A.~Vedaldi, and A.~Zisserman, ``Deep inside convolutional
  networks: Visualising image classification models and saliency maps,''
  \emph{arXiv preprint arXiv:1312.6034}, 2013.

\bibitem{gbp}
J.~T. Springenberg, A.~Dosovitskiy, T.~Brox, and M.~Riedmiller, ``Striving for
  simplicity: The all convolutional net,'' \emph{arXiv preprint
  arXiv:1412.6806}, 2014.

\bibitem{cam}
B.~Zhou, A.~Khosla, A.~Lapedriza, A.~Oliva, and A.~Torralba, ``Learning deep
  features for discriminative localization,'' in \emph{Proceedings of the IEEE
  Conference on Computer Vision and Pattern Recognition}, 2016, pp. 2921--2929.

\bibitem{grad-cam}
R.~R. Selvaraju, M.~Cogswell, A.~Das, R.~Vedantam, D.~Parikh, and D.~Batra,
  ``Grad-cam: Visual explanations from deep networks via gradient-based
  localization,'' in \emph{Proceedings of the IEEE International Conference on
  Computer Vision}, 2017, pp. 618--626.

\bibitem{grad-campp}
A.~Chattopadhay, A.~Sarkar, P.~Howlader, and V.~N. Balasubramanian,
  ``Grad-cam++: Generalized gradient-based visual explanations for deep
  convolutional networks,'' in \emph{2018 IEEE Winter Conference on
  Applications of Computer Vision (WACV)}.\hskip 1em plus 0.5em minus
  0.4em\relax IEEE, 2018, pp. 839--847.

\bibitem{score-cam}
H.~Wang, Z.~Wang, M.~Du, F.~Yang, Z.~Zhang, S.~Ding, P.~Mardziel, and X.~Hu,
  ``Score-cam: Score-weighted visual explanations for convolutional neural
  networks,'' in \emph{Proceedings of the IEEE/CVF Conference on Computer
  Vision and Pattern Recognition Workshops}, 2020, pp. 24--25.

\bibitem{ss-cam}
H.~Wang, R.~Naidu, J.~Michael, and S.~S. Kundu, ``Ss-cam: Smoothed score-cam
  for sharper visual feature localization,'' \emph{arXiv preprint
  arXiv:2006.14255}, 2020.

\bibitem{kd}
G.~Hinton, O.~Vinyals, and J.~Dean, ``Distilling the knowledge in a neural
  network,'' \emph{arXiv preprint arXiv:1503.02531}, 2015.

\bibitem{kd2}
J.~Ba and R.~Caruana, ``Do deep nets really need to be deep?'' \emph{Advances
  in Neural Information Processing Systems}, vol.~27, 2014.

\bibitem{kd3}
A.~Romero, N.~Ballas, S.~E. Kahou, A.~Chassang, C.~Gatta, and Y.~Bengio,
  ``Fitnets: Hints for thin deep nets,'' \emph{arXiv preprint arXiv:1412.6550},
  2014.

\bibitem{kd4}
N.~Komodakis and S.~Zagoruyko, ``Paying more attention to attention: improving
  the performance of convolutional neural networks via attention transfer,'' in
  \emph{ICLR}, 2017.

\bibitem{kd5}
J.~Yim, D.~Joo, J.~Bae, and J.~Kim, ``A gift from knowledge distillation: Fast
  optimization, network minimization and transfer learning,'' in
  \emph{Proceedings of the IEEE Conference on Computer Vision and Pattern
  Recognition}, 2017, pp. 4133--4141.

\bibitem{kd6}
X.~Wang, R.~Zhang, Y.~Sun, and J.~Qi, ``Kdgan: Knowledge distillation with
  generative adversarial networks,'' \emph{Advances in Neural Information
  Processing Systems}, vol.~31, 2018.

\bibitem{kd7}
H.~Lin, G.~Han, J.~Ma, S.~Huang, X.~Lin, and S.-F. Chang, ``Supervised masked
  knowledge distillation for few-shot transformers,'' in \emph{Proceedings of
  the IEEE/CVF Conference on Computer Vision and Pattern Recognition}, 2023,
  pp. 19\,649--19\,659.

\bibitem{auc}
P.~Flach, J.~Hern\'{a}ndez-Orallo, and C.~Ferri, ``A coherent interpretation of
  auc as a measure of aggregated classification performance,'' in
  \emph{Proceedings of the 28th International Conference on International
  Conference on Machine Learning}, 2011, p. 657–664.

\bibitem{roc}
``The use of the area under the roc curve in the evaluation of machine learning
  algorithms,'' \emph{Pattern Recognition}, vol.~30, no.~7, pp. 1145--1159,
  1997.

\bibitem{box}
D.~Manolakis and G.~Shaw, ``Detection algorithms for hyperspectral imaging
  applications,'' \emph{IEEE Signal Processing Magazine}, vol.~19, no.~1, pp.
  29--43, 2002.

\bibitem{adam}
D.~P. Kingma and J.~Ba, ``Adam: A method for stochastic optimization,''
  \emph{arXiv preprint arXiv:1412.6980}, 2014.

\bibitem{ema}
A.~Tarvainen and H.~Valpola, ``Mean teachers are better role models:
  Weight-averaged consistency targets improve semi-supervised deep learning
  results,'' in \emph{Advances in Neural Information Processing Systems},
  I.~Guyon, U.~V. Luxburg, S.~Bengio, H.~Wallach, R.~Fergus, S.~Vishwanathan,
  and R.~Garnett, Eds., vol.~30.\hskip 1em plus 0.5em minus 0.4em\relax Curran
  Associates, Inc., 2017.

\end{thebibliography}
\bibliographystyle{IEEEtran}

\end{document}